\documentclass{article}
\usepackage{graphicx} 
\usepackage{float}
\usepackage{subcaption}

\usepackage[english]{babel} 
\usepackage[utf8]{inputenc} 
\usepackage{csquotes}
\usepackage{amsmath}

\usepackage[round, sort&compress, authoryear]{natbib}
\usepackage[colorlinks=true,   
            linkcolor=red,     
            urlcolor=magenta,  
            citecolor=blue    
           ]{hyperref}
           
\usepackage[final]{neurips_2024}

\usepackage[T1]{fontenc}    
\usepackage{hyperref}       
\usepackage{url}            
\usepackage{booktabs}       
\usepackage{amsfonts}       
\usepackage{nicefrac}       
\usepackage{microtype}      
\usepackage{xcolor}         

\title{Aquarius: A Family of Industry-Level Video Generation Models for Marketing Scenarios}

\author{%
  Huafeng Shi \textsuperscript{*}\\
  \And
  Jianzhong Liang \textsuperscript{*} \\
  \And
  Rongchang Xie  \\
  \And
  Xian Wu \\
  \And
  Cheng Chen \\
  \And
  Chang Liu \\
  ByteDance \\
}

\begin{document}
\maketitle

\begin{abstract}
This report introduces \textbf{Aquarius}, a family of industry-level video
generation models for marketing scenarios designed for thousands-xPU clusters and models with hundreds of billions of parameters. Leveraging efficient engineering architecture and algorithmic innovation, Aquarius demonstrates exceptional performance in high-fidelity, multi-aspect-ratio, and long-duration video synthesis. By disclosing the framework's design details, we aim to demystify industrial-scale video generation systems and catalyze advancements in the generative video community. The Aquarius framework consists of five components: \textit{\textbf{Distributed Graph and Video Data Processing Pipeline:}} Manages tens of thousands of CPUs and thousands of xPUs via automated task distribution,  enabling efficient video data processing. Additionally, we are about to open-source the entire data processing framework named "\textit{Aquarius-Datapipe}". \textit{\textbf{Model Architectures for Different Scales:}} Include a Single-DiT architecture for 2B models and a Multimodal-DiT architecture for 13.4B models, supporting multi-aspect ratios, multi-resolution, and multi-duration video generation. \textit{\textbf{High-Performance infrastructure designed for video generation model training:}} Incorporating hybrid parallelism and fine-grained memory optimization strategies, this infrastructure achieves 36\% MFU at large scale. \textit{\textbf{Multi-xPU Parallel Inference Acceleration:}} Utilizes diffusion cache and attention optimization to achieve a 2.35x inference speedup. \textit{\textbf{Multiple marketing-scenarios applications:}} Including image-to-video, text-to-video (avatar), video inpainting and video personalization, among others. More downstream applications and multi-dimensional evaluation metrics will be added in the upcoming version updates.
\end{abstract}

\section{Introduction}
The release of OpenAI's Sora \citep{videoworldsimulators2024} video generation model in 2024 marked a paradigm shift in video generation, demonstrating superior performance compared to traditional generative adversarial network \citep{goodfellow2014generativeadversarialnetworks} (GAN) approaches and igniting widespread research interest in video generation technologies. Enabled by open-source community contributions, video generation technology has achieved remarkable progress over the past year. By mid-late 2024, pioneering projects including Open-Sora \citep{peng2025opensora20trainingcommerciallevel}, Open-Sora-Plan \citep{lin2024opensoraplanopensourcelarge}, EasyAnimate \citep{xu2024easyanimatehighperformancelongvideo} and CogVideoX \citep{yang2025cogvideoxtexttovideodiffusionmodels} had open-sourced their training frameworks and model weights, marking the first public disclosure of implementation details in video generation. Nevertheless, constrained by the limited data scale, model architectures, and computational resources of development teams, these open-source solutions initially exhibited significant performance gaps compared to proprietary commercial models. With substantial resource investments from leading open-source research teams, the effects demonstrated by video generation works such as HunyuanVideo \citep{kong2025hunyuanvideosystematicframeworklarge} and WanVideo \citep{wan2025wanopenadvancedlargescale} have significantly narrowed the gap between open-source models and commercial counterparts \citep{Runway, Luma}. This advancement has further driven the development of the generative AI community and confirmed the effectiveness of diffusion paradigms in video generation capabilities.

DiT \citep{peebles2023scalablediffusionmodelstransformers} has validated that the Transformer-based structure can better realize the scaling law in diffusion tasks. Contemporary video generation models predominantly employ the Video Diffusion Transformer (VideoDiT) architecture, which achieves unified capabilities for image and video generation through tokenization of visual information. Visual generation frameworks typically incorporate two modalities: textual modality and visual modality. Based on the interaction methods of these dual modalities, we categorize current visual generation frameworks into two paradigms.

\textbf{Single-DiT.} In the Single-DiT architecture, the visual branch accounts for the majority of the model's parameters and computational resources. The textual modality through dedicated encoders extract features and then interacts with the visual modality via cross-attention layer.   The advantages of this architecture is algorithmic simplicity and straightforward engineering implementation, enabling prioritized allocation of computational resources to the visually critical components for enhanced video generation quality. Key limitations include restricted multimodal expansion capabilities and suboptimal text comprehension performance.  Representative image generation models include SD1.5 \citep{rombach2022highresolutionimagesynthesislatent}, SDXL \citep{podell2023sdxlimprovinglatentdiffusion}, Pixart-Alpha \citep{chen2023pixartalphafasttrainingdiffusion}, and DiT, while representative video generation models include Open-Sora V1.0, Open-Sora-Plan V1.2 and WanVideo. We utilized the Single-DiT framework under the 2B model, with the generation results as shown in the Figure \ref{fig:attribute}. It shows a good instruction-following ability and fine-grained video generation capabilities.

\textbf{Multimodal-DiT.} In the multi-stream framework, visual and textual information are processed through separate branches to learn their own modality-specific features. In the early stages of the framework, each layer conducts preliminary information interaction via self-attention. In the later stages, each layer performs in-depth information interaction by concatenating visual tokens and text tokens and applying self-attention. Finally, only the visual modality is separated for video generation. The advantages of this architecture include strong modality extension capabilities (e.g., audio) and enhanced multimodal modeling abilities. The main limitations are its complexity in implementation and the potential for redundant computation in non-visual modalities, which may affect and diminish the learning capability of the visual modality. Representative image generation models include SD3 \citep{esser2024scalingrectifiedflowtransformers} and Flux \citep{flux2024}, while representative video generation models include CogVideoX, HunyuanVideo, and Open-Sora V2.0. We utilized the Multimodal-DiT framework under the 13.4B model, with the generation results as shown in the Figure \ref{fig:digitalman_2}.

The success of HunyuanVideo (13B) and WanVideo (14B) has demonstrated that further scaling up model parameters under the VideoDiT architecture can lead to significant performance improvements. However, for most teams, training a commercial-grade video generation model with billions of parameters from scratch remains a formidable challenge. On one hand, current technological reports tend to focus on introducing coarse-grained video generation frameworks and showcasing their effects, while lacking detailed explanations of critical training design elements. This results in a high trial-and-error cost even when training code is available. On the other hand, there is a lack of experience in constructing efficient and stable training frameworks on a thousand-card xPUs scale, which directly leads to inefficient training, resource wastage, and even the inability to load training models. This paper mainly addresses and publicizes the aforementioned issues to further promote the continuous growth and innovation of the open-source video generation community.

The structure of this technical report is as follows:
\begin{itemize}
\item In\textbf{ Section \ref{section2}}, we introduce how to build a large-scale data processing pipeline based on a distributed framework, including data filtering, data labeling, and data refinement.
\item In\textbf{ Section \ref{section3}}, we present the architectural design details and design principles of the Aquarius 2B\&13.4B video generation models.
\item In\textbf{ Section \ref{section4}}, we discuss the model training strategies and considerations.
\item In\textbf{ Section \ref{section5}}, we describe how to construct a stable and efficient infra architecture at the scale of thousands of xPUs.
\item In\textbf{ Section \ref{section6}}, we describe how to efficiently perform parallel accelerated inference for video generation on multiple xPUs.
\item In\textbf{ Section \ref{section7}}, we demonstrate a variety of application models in marketing scenarios, such as digitalman generation, video inpainting and video personalization, etc.
\item In\textbf{ Section \ref{section8}} and\textbf{ \ref{section9}}, we summarize the technical report and discuss future exploratory work.
\end{itemize}

\section{Data Curation}
\label{section2}
In the era of large models, abundant and high-quality training data is paramount.  Due to differences in media formats, video data presents challenges several orders of magnitude greater than text or image data in terms of processing efficiency, computational complexity, and storage requirements. This makes an efficient and flexible data curation framework particularly crucial for video generation tasks. Our data processing framework, Aquarius-Datapipe, is developed based on the Ray \citep{moritz2018raydistributedframeworkemerging} architecture. Ray is an open-source framework for building and running distributed applications, specifically designed to simplify distributed computing for machine learning workloads while supporting complex task graphs.  This enables developers to create parallel or distributed programs that efficiently leverage all available cluster resources. Aquarius-Datapipe integrates dozens of image and video processing operators, allowing rapid construction of video filtering workflows through configuration files.  It also enables precise allocation of computing devices for specific processing operators.  While Aquarius supports both image and video training/generation through its DiT framework, image processing can be regarded as a subtask of video processing. This section focuses on describing the video processing workflow.

\subsection{Data Pre-processing}
Given the substantial computational resource demands of video processing, we adhere to the following design principles when configuring video data pipelines:
\begin{itemize}
\item Prioritize operators with stricter filters (smaller "funnel apertures") to reduce the processing load for subsequent operators.
\item Prioritize operators that are processed by the CPU and have lower CPU usage to save xPU resources.
\item Prioritize computationally efficient operators to maximize data throughput.
\end{itemize}

\subsection{Data Filtering}
\subsubsection{Video-level Data Filtering}
We prioritize filtering out video files with abnormal aspect ratios, low frame rates (fps), low sampling rates, and short durations by examining video metadata. We also randomly sample frames from videos to detect and filter out unwanted elements such as black borders, titles, and subtitles. Additionally, we conduct preliminary diagnostics on video files to ensure they will not cause issues during subsequent processing. These operations only require CPU resources.
\subsubsection{Clip-level Data Filtering}
\textbf{Video cropping.} The PySceneDetect \citep{PySceneDetect} algorithm relies on frame-to-frame differences is unable to handle gradual transitions, leading to scene-jumping issues in model generation. To address this, we trained our own scene detection model based on TransNetV2 \citep{soucek2020transnetv2} using internal data. Empirical validation demonstrates superior performance in handling complex scene transitions, including: smooth crossfades and dissolves, flash effects and motion blur transitions. We use the crop-detect method in FFmpeg \citep{Ffmpeg} to detect and crop the black borders of the video clips after splitting. Then, we employ an OCR detection model to identify and crop the subtitles at the top and bottom of the videos. Finally, we filter out videos with an excessive amount of text.

\textbf{Clip quality filtering.} We employ a variety of strategies to ensure the aesthetic quality and safety of pre-training videos. \textbf{(1) Content Security}: Our data sources have all undergone category labeling and NSFW (Not Safe For Work) filtering. Therefore, we primarily balance the data types based on label information. \textbf{(2) Visual quality filtering}: we discovered that the aesthetic scoring model has relatively low accuracy for videos of medium quality, we only use the scores filtered out videos with abnormally low scores. Then, we used internally trained multi-dimensional video quality model to evaluate the visual aesthetics of video clips from both aesthetic and technical viewpoints. Finally, we used our internal human quality model to score and filter based on human integrity, facial integrity, attractiveness, clarity, and whether there is any occlusion. It is important to note that the threshold values of the above filtering models need to be readjusted for different resolution stages. \textbf{(3) Clip deduplication}: We use our internal VideoCLIP model to extract features and remove duplicates from the video clips. Additionally, we use clustering to assist with label information for data classification balance. \textbf{(4) Motion filtering}: We use optical flow to predict the motion magnitude of videos and filter out clips with low or excessively high motion.

\subsection{Data Post-processing}
In the post-training phase, the primary objective is to significantly enhance the fidelity and motion quality of generated videos based on high-quality data.  This part is mainly achieved through both human evaluation and manual collection.Specifically, we further refine the data by improving the quality score to obtain a smaller dataset.  The former involves human review, where reviewers score and filter videos based on the completeness of motion, authenticity of image quality, subjective aesthetic appeal, and integrity of the video subject.  The latter primarily involves acquiring high-aesthetic, cinematographic-level video data through manual filming and post-production processes.

\begin{figure}
    \centering
    \includegraphics[width=1\linewidth]{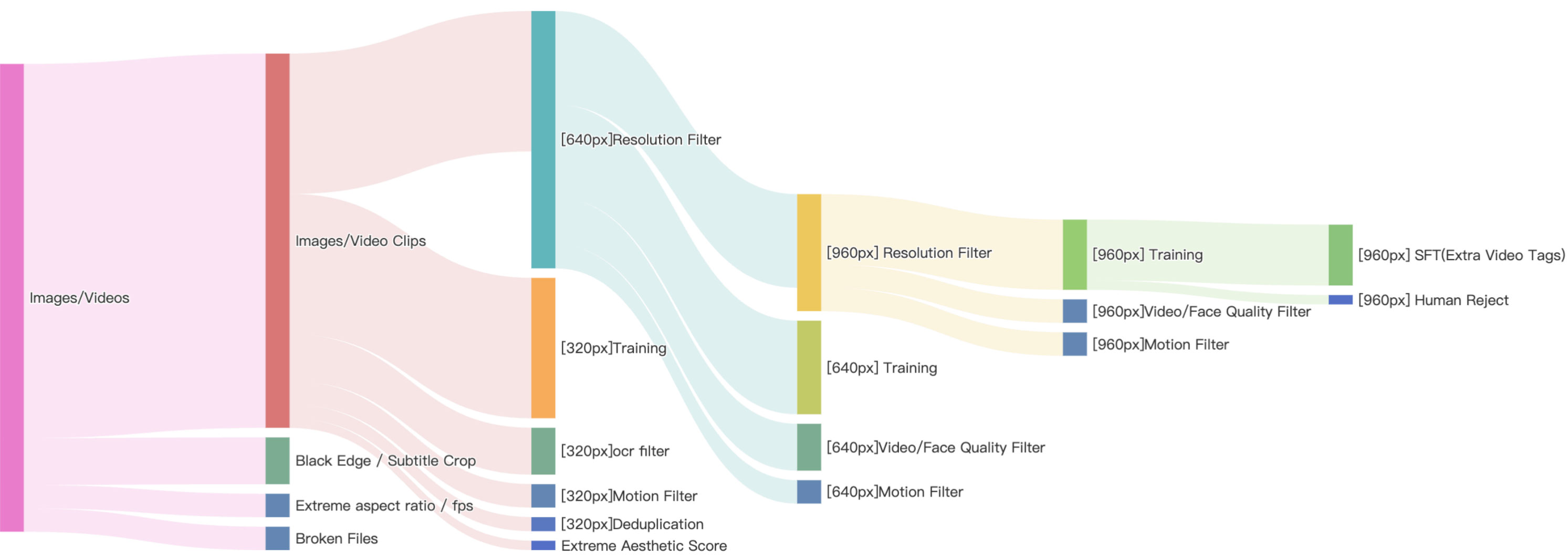}
    \caption{Data Filter Pipeline}
    \label{fig:enter-label}
\end{figure}

\subsection{Data Annotation}
Using high-quality caption models for billion-level image and video data is extremely demanding on xPU resources. To address this, we employ in-house trained, lighter video captioning models with FP8 quantization and a parameter size of 7B to process large volumes of low-quality pre-training data. For high-quality data, we utilize an in-house trained video captioning model with 34B parameters and superior performance to process the data.

\subsubsection{Structured Captioning}
We have designed prompts to guide video caption models in generating both video summaries and structured summaries. The video summaries are used to briefly outline the motion state of the subject. The structured summaries provide detailed descriptions of the video: \textbf{(1)} Scene composition. \textbf{(2)} Subject appearance details. \textbf{(3)} Motion characteristics of subjects. \textbf{(4)} Background environment. \textbf{(5)} Visual style (e.g., cinematic, documentary). \textbf{(6)} Camera techniques (shot types, angles, movements. \textbf{(7)} Lighting conditions. The structured description approach can provide an effective template for LLM model to rewrite user prompts, facilitating the alignment of inference prompts.

\subsubsection{Extra Captioning}
We refer to Open-Sora and concatenate information such as video quality scores, camera motion, and face quality score into the caption. We find that the model has a certain degree of adherence to quality scores and camera control, especially in the early stages of model training. This adherence gradually diminishes in the later stages of training.

\section{Model Architecture}
\label{section3}
\subsection{3D Variational Autoencoder}
Variational Autoencoders (VAEs) accelerate the training speed of diffusion models and reduce training resources by compressing high-dimensional visual data into a compact latent space. However, videos pose two unique challenges in VAE encoding compared to images: \textbf{(1)} Videos contain both visual and motion information, making joint spatial-temporal modeling particularly crucial. \textbf{(2)} The larger input size of videos can lead to xPU memory overflow during the encoding and decoding inference stages.

\begin{figure}
    \centering
    \includegraphics[width=1\linewidth]{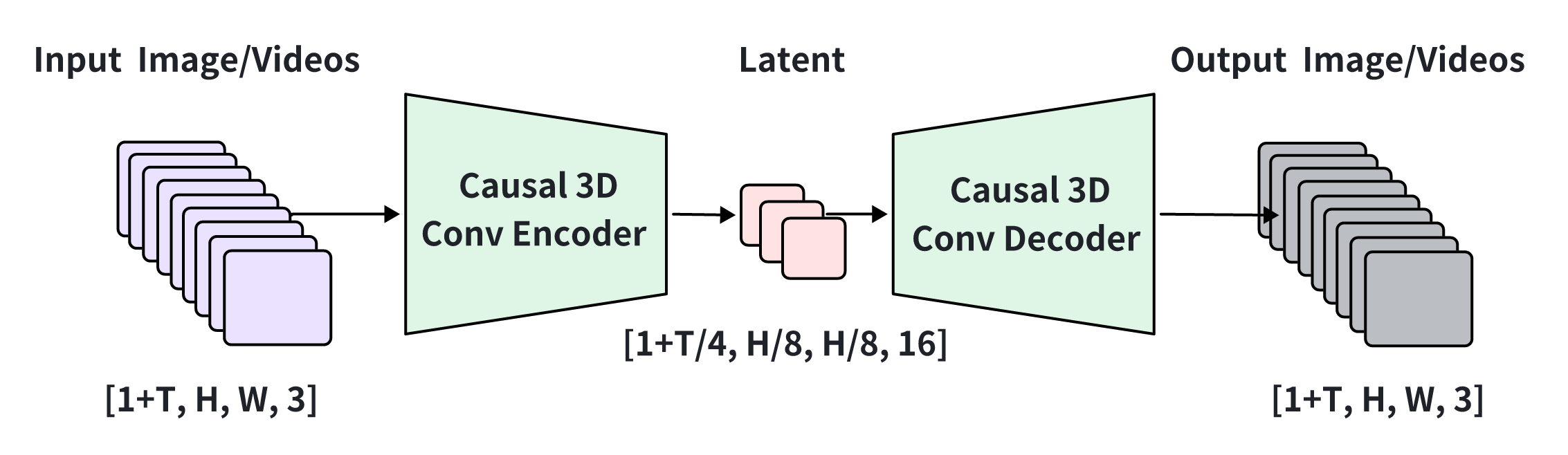}
    \caption{Overview of our 3DVAE.}
    \label{fig:3DVAE}
\end{figure}

We leverages CausalConv3D \citep{yu2024languagemodelbeatsdiffusion} to achieve unified spatial-temporal modeling across images and videos, where images are treated as a special case with temporal dimension T=1. This design enables consistent feature learning while maintaining generation quality across modalities. As shown in the Figure \ref{fig:3DVAE}, the 3D VAE consists of an encoder and a decoder. Specifically, the encoder compresses an input video of shape $(1 + T_{in}) \times H_{in} \times W_{in}$ into a compact representation of shape $T_{out}\times H_{out} \times W_{out} = (1 + T_{in}/4) \times H_{in}/8\times w_{out}/8$, with a compression ratio of $4\times8\times8$. The decoder then upsamples this compact representation back to the original video shape of $(1 + T_{in}) \times H_{in} \times W_{in}$, either from the encoder's output or from the output of a DiT model.

We perform spatial-temporal tiling strategy to divide video into blocks in both temporal and spatial dimensions and then blend the corresponding block features through a linear combination to obtain the final video features. During training, we adopt a multi-stage training paradigm similar to DiT (as illustrated in the Table \ref{tab:training_details}). Unlike DiT, our approach employs a tiling strategy, which means that the training frames and resolution do not need to be as large as those in DiT to achieve satisfactory extrapolation results. The tiling approach has two main advantages. On one hand, it reduces the computational and memory requirements for both training and inference while supporting extrapolation of videos of arbitrary length. On the other hand, it facilitates parallelization across multiple xPUs by assigning different tiles to different xPUs, thereby achieving linear-level acceleration.

To further improve the compression ratio while maintaining comparable reconstruction quality, a common approach is to increase the number of output latents feature channels to compensate for information loss. In our experiments, we found that larger channel dimensions in the feature representation pose greater challenges for subsequent diffusion model training. This is because the feature channel information is only involved in a small amount of computation during the initial stages of DiT, and it further increases the difficulty of the training objective.

\subsection{Dit Model Architecture}
\begin{figure}
    \centering
    \includegraphics[width=1\linewidth]{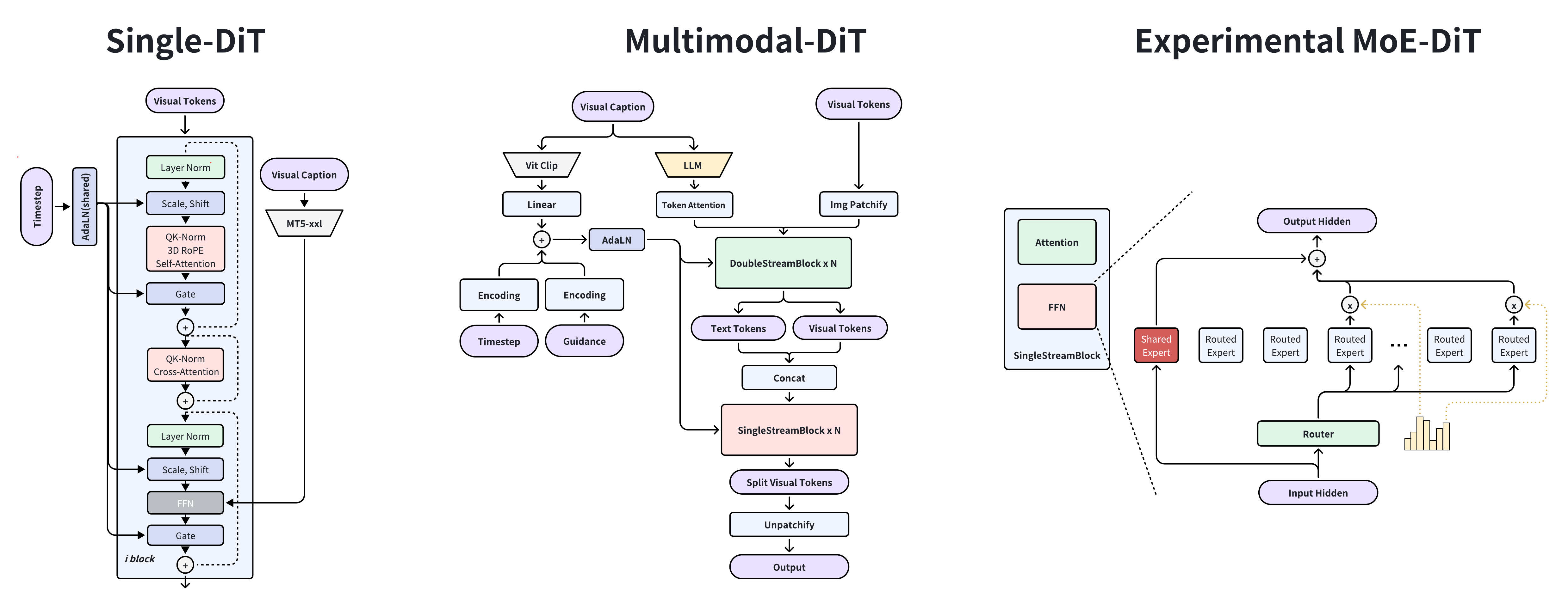}
    \caption{Overview of the our VideoDiT architecture.}
    \label{fig:model architecture}
\end{figure}

We present distinct DiT architectures tailored for different model scales: a compact 2B parameter model based on single-stream network architecture and a larger 13.4B parameter model employing dual-stream network architecture. Text-to-video generation task typically involves visual and textual modalities, with the visual modality being significantly more challenging to train than the textual modality. Textual inputs leverage pre-trained high-performance text models to extract semantically dense representations while requiring relatively simple processing, we posit that prioritizing computational allocation to the visual modality under constrained parameter budgets yields optimal performance. Conversely, when sufficient computational resources are available, expanding capacity for textual modality processing enhances semantic reasoning and cross-modal alignment capabilities. This analysis leads to our architectural strategy: implementing single-stream architecture for 2B models and adopting multimodal architecture for 13.4B or larger-scale models. At the same time, we have further explored a scheme for verifying the scaling law based on Mixture of Experts \citep{moe} (MoE), and conducted validation tests during the multi-stage training of the text-to-image phase. The detailed model architecture is illustrated in Figure \ref{fig:model architecture}.

\subsubsection{Single DiT}
Aquarius2B-single DiT, we employ the mT5-XXL \citep{xue2021mt5massivelymultilingualpretrained} as the text encoder, which supports both Chinese and English understanding and provides robust text comprehension capabilities, thereby further reducing the difficulty of learning in the text modality. Following AdaLN-single approach from Pixart-Alpha, we share the modulation module weights to further reduce model parameters. The text modality interacts with visual tokens through cross-attention at each block of DiT. Finally, the output visual tokens are transformed back to the latent space via an unpatchify operation and then decoded into the final video through the 3D VAE decoder. 

\subsubsection{Multimodal DiT}
Aquarius13.4B-multimodal DiT, we employ an internally trained Multimodal Large Language Model (MLLM) as the text encoder to extract features. This model not only possesses enhanced capabilities for fine-grained text understanding and visual-text alignment but also incorporates the reasoning abilities of a Large Language Model (LLM), greatly improve the zero-shot capability for video generation. To further enhance the guiding capability of text features, we employ the CLIP \citep{radford2021learningtransferablevisualmodels} model to extract these features. These features are mapped and added to the timestep features to jointly guide video generation. We utilize the AdaLN-zero modulation model without weight sharing and further partition the parameters through a TP/SP (Tensor Parallelism/Sequence Parallelism) strategy to significantly reduce the parameter count.

\subsubsection{Single MoE-DiT}
Aquarius 10BA1.7B MoE-DiT, we build an experimentally sparse activation model to explore the performance and efficiency of MoE-based diffusion Transformers in image generation. Building on successful MoE practices \citep{dai2024deepseekmoe} from Large language models (LLMs), we substitute a subset of the dense-FFNs with MoE-FFNs in SingleStreamBlocks of the model. Experts are segmented into finer-grain with intermediate hidden dimension reduced to 1/4 of the original FFN layer. In each MoE layer, two shared experts are permanently activated to capture common knowledge, while 6 of 64 routered experts are dynamically activated to enhance diversity and specialization. The experimental model comprises 10B total parameters with 1.7B parameters activated per forward pass. We adopt expert parallelism to distribute routed experts in each MoE layer across 8 xPUs within a single node, and apply pipeline parallelism to partition the model into multiple transformer-layer-based stages across nodes.

\textbf{3D Full Attention.}
In past experiments, spatial-temporal separated attention mechanisms have often been shown temporal inconsistency and subject distortion. Open-Sora-Plan V1.3 propose the Skiparse (Skip-Sparse) Attention method. Specifically, under a fixed sparse ratio k and candidate tokens for attention through two alternating skip-gather methods. This approach preserves the attention operation is global while effectively reducing FLOPS, enabling faster training of 3D Attention models. We implemented Full Attention and Sparse Attention by setting K to 1 and 4, respectively, at different stages of model training to reduce computational load. However, we found that using full attention throughout the entire training process consistently yielded more stable video generation results. Ultimately, we decided to use full attention for end-to-end training in both model sizes.

\textbf{3D RoPE.}
To enable a single modality to support multi-resolution, multi-aspect-ratio, and multi-frame-length video generation, we apply 3D RoPE encoding. We apply 1D Rotary Position Embedding \citep{su2023roformerenhancedtransformerrotary} (RoPE) encoding separately to the temporal, height, and width dimensions, assigning different channel numbers to each. The features are then combined and concatenated to form the attention head dimension size. Given the separate encoding of the three dimensions, we choose a lower frequency base to enhance the sensitivity of the position encoding to the temporal and spatial details of the video.

\section{Model Training}
\label{section4}
\subsection{Dataset Strategy}
We employ a bucket-based strategy to construct video dataset where each bucket of video data is characterized by a quadruple $\{batchsize, frames, height, width \}$. During training, multiple bucket configurations can be dynamically allocated. For instance, given a training video with dimensions $(T_{s}\times H_{s} \times W_{s})$, our method first identifies the nearest bucket frame length exceeding $T_{s}$ and random temporal crop the video to the bucket's specified frame length. Subsequently, based on the product of $H_{s}$ and $W_{s}$, we find the closest bucket value and resize the video's height and width to match the bucket's width and height. Finally, by adjusting the batchsize dimension, we optimize the total token count to maximize xPU resource utilization while avoiding xPU memory overflow and reducing the waiting time for tasks across different xPUs within the same batch.

Specifically, video configurations such as $$\{1,29,640,640\},\{1,29,480,854\},\{1,29,854,480\},\{1,125,320, 320\},\{8,29,320, 320\}$$, all have the same number of tokens and can be trained efficiently in DP(data parallel). This strategy conveniently enables the construction of a video generation training dataset with multiple resolutions, aspect ratios, and frame lengths. By simply modifying the configuration file, we can complete the entire video generation training process, ultimately achieving mixed-scale video generation with the aid of 3D-RoPE encoding.

\subsection{Training Strategy}
We adopt the Flow Matching \citep{lipman2023flowmatchinggenerativemodeling} framework to model across both image and video domains. Compared to Denoising Diffusion Probabilistic Models \citep{ho2020denoisingdiffusionprobabilisticmodels} (DDPM) , Flow Matching provides a more direct and efficient way to model the data distribution, enabling rapid convergence to high-quality outputs when training from scratch and significantly reducing the number of inference sampling steps.
\subsubsection{Training Objective} 
In Flow Matching, the goal is to learn a function that can transform a simple base distribution (such as a Gaussian) into the target data distribution by following this trajectory. This is achieved by minimizing the difference between the flow of the data and the flow predicted by the model. We first sample $X_{0}$ from a Gaussian noise distribution  $X_{0} \sim \mathcal{N}(0,1)$, sample $t \in [0, 1]$ from logit-normal distribution, and obtain intermediate latent $X_t$ via linear interpolation as $X_t = (1 - t) \cdot X_0 + t \cdot X_1$.
The ground truth is the direction and magnitude of the velocity change from the initial noise $X_0$ to the target video $X_1$:
\begin{equation}
V_t = \frac{dX_t}{dt} = X_1 - X_0.
\end{equation}
The DiT model predicts the flow velocity based on the current timestep $t$, intermediate latent $X_t$ and the condition $y$. The loss is represented as the mean squared error between the model's predicted velocity and the true velocity.

\begin{equation}
\mathcal{L} = \mathbb{E}_{t, x_0, x_1, y} \left\| u(X_t, y, t; \theta) - V_t \right\|^2,
\end{equation}

\subsubsection{Multi-Stage Strategy} 
Our empirical observations reveal that motion patterns exhibit significantly slower learning convergence compared to visual fidelity during resolution-duration scaling. Specifically, when transitioning from low-resolution, short-duration training phases to high-resolution, long-duration regimes, visual quality typically converges to target benchmarks within several thousand iterations. In contrast, temporal coherence requires extended training periods (10k+ steps) and may never reach the desired level for the current stage. Guided by curriculum learning principles and extensive ablation studies, we have carefully designed multiple training stages in the hope of achieving better results with lower training resources, as shown in the Table \ref{tab:training_details}. Notably, all multi-stage training can be accomplished simply by modifying the bucket dataset configuration files.

\textbf{Text-to-image Pre-training.} The text-to-image pre-training stage aims to quickly establish a connection between text descriptions and visual information during the cold start phase.   There are two considerations for expanding the resolution from $320px$ to $640px$ in this stage.   First, we found that models trained at higher resolutions exhibit better transfer effects when switched to lower-resolution models, thereby providing a better initialization model for the $320px$ video training stage.   Second, after correctly training the model at the $640px$ resolution, it is sufficient to obtain an image generation model with reasonably good visual effects.   We can quickly verify any potential bugs in the model training framework at this stage.   This strategy allows us to use extremely low resources to identify and eliminate over 90\% of hidden bugs in the Aquarius architecture, significantly shortening the model debugging process.

\textbf{Text-to-video Pre-training.} The text-to-video pre-training stage focuses on establishing cross-modal alignment between textual descriptions and spatiotemporal dynamics, constituting a critical expansion from static image generation to temporal-aware video synthesis.  In this stage, we expand the number of frames from 29 to 61 at 320px resolution. Given the relatively low training resource requirements, we recommend training for a longer number of steps, approximately two epochs. We expect to obtain a model with decent motion modality at 320px resolution by the end of this stage. Similarly, this stage also helps us identify and eliminate bugs in the Aquarius architecture related to temporal processing.

\textbf{Text-to-image / video Joint Pre-training.} The text-to-image / video joint pre-training stage focuses on dual enhancement of visual fidelity and motion consistency in video generation. We achieve this by modifying the data bucket configuration file to conduct joint training of single-frame images and multi-frame videos in the same training process.     Experimental results demonstrated that the appropriate introduction of image supervision could further strengthen the instruction-following capability while facilitating rapid convergence of videos to the desired high-resolution standards. This effect can be further enhanced by setting unequal resolutions for images and videos.   Notably, excessive use of images was observed to impair motion information learning, leading to sluggish motion in generated videos. To mitigate this, we progressively reduced the image ratio from 30\% to 10\% throughout the training process, balancing visual fidelity with motion consistency.

\textbf{Supervised Fine-tuning.} This SFT stage aims to further enhance visual quality to align with human preferences and specific downstream application scenarios.   We implemented more stringent filtering criteria on the data from the previous stage and meticulously curated a high-quality dataset of approximately 500k samples through manual shooting and planning.   This dataset focuses on more challenging human-centric scenes, featuring individuals with high aesthetic appeal, dynamic movements, and a sense of realism. Significant improvements in body posture coordination and realism were achieved after the supervised fine-tuning (SFT) phase, particularly in complex human interactions and movements.

\textbf{Early Stop.} In later training stages, the model's increased capability combined with a reduction in data volume can lead to overfitting, particularly during the SFT stage.  While this issue can be somewhat mitigated by employing model Exponential Moving Average (EMA) strategies, it can still impact the quality of the generated output. To prevent model performance degradation, we periodically generate videos at set training intervals for manual inspection and evaluation. Empirically, training for 1.5 epochs tends to yield satisfactory results.

\begin{table}[htbp]
    \centering
    \begin{tabular}{p{25mm}p{25mm}p{25mm}p{15mm}p{15mm}p{15mm}}
        \toprule [1pt]     
        Training Stage & Image Resolution & Video Resolution & Batchsize & Step & Adam Lr\\
        \midrule [1pt]
        Stage1: T2I & 1x320x320 & - & 4096 & O(100K) & 1e-4 \\
        & 1x640x640 & - & 2048 & O(100K) & 1e-4 \\
        Stage2: T2V & - & 29x320x320 & 1024 & O(100K) & 1e-4 \\
        & - & 61x320x320 & 1024 & O(100K) & 1e-4 \\
        Stage3: T2I/V & 1x640x640 & 61x640x640 & 1024 & O(100K) & 1e-4 \\
        & 1x960x960 & 125x640x640 & 512 & O(10K) & 4e-5 \\
        & 1x960x960 & 125x960x960 & 256 & O(10K) & 4e-5 \\
        SFT: T2I/V & 1x960x960 & 125x960x960 & 128 & O(1K) & 1e-5 \\
        & 1x1440x1440 & 125x960x960 & 128 & O(1K) & 1e-5 \\
        \bottomrule [1pt]     
    \end{tabular}
    \caption{Multi-Stage Training details.}
    \vspace{-2mm}
    \label{tab:training_details}
\end{table}

\section{Training Efficieny}
\label{section5}
Aquarius's training is both memory-intensive and compute-intensive. Compared to general LLM training practices \citep{narayanan2021efficient, grattafiori2024llama}, Aquarius's structural variations and extended sequence lengths require tailored optimizations in training strategies to enhance performance. Notably, parameters introduced by modules like patchify and adaLN demand specialized handling under tensor parallelism. Significant sequence length variations across training stages alter memory consumption and computation-to-communication patterns, necessitating adaptive parallelism adjustments combined with fine-grained control over recomputation and offloading.

Training is conducted on a cluster comprising thousands of accelerators. During the most computationally intensive stage of training, our 13.4B model achieves a Model FLOPs Utilization (MFU) of 36\%. This section details our observations and practical experiences from optimizing the 13.4B Aquarius model's training efficiency.

\textbf{Decouple Encoders and DiT.} The complete training workflow of Aquarius involves a VAE model as video encoder, two distinct text encoder models, and a backbone DiT model. During training, all encoder parameters remain frozen, with their outputs serving as the input sequence for DiT Training. Concurrently loading all four models (VAE, dual text encoders, DiT) into same devices during training is not efficient in our experiments :
\begin{itemize}
    \item \textbf{Memory inefficiency:} Due to long video sequences, memory consumption during DiT training is extremely high. Non-trainable parameters of the frozen encoders would compete for limited memory resources, further aggravating the memory bottleneck.
    \item \textbf{Gradient accumulation constraint:} The optimal parallelism strategies differ between DiT and encoders, with the encoders employing a larger Data Parallelism (DP) size. To prevent data inefficiency caused by mismatched rates between the encoder data production and the DiT data consumption, the gradient accumulation steps in DiT training must equal $\frac{encoder\_dp\_size}{dit\_dp\_size}$, which restricts the flexibility of this hyperparameter.
\end{itemize}
Therefore, we decoupled the encoder inference and DiT training phases into separate tasks. Data is first generated by the encoder inference task and subsequently consumed by the training task. This decoupled paradigm also enables us to allocate optimized computational resources with distinct specifications for DiT training and encoder inference phases.

\textbf{Memory Consumption.}  Prior to implementing specific training optimizations, we estimate xPU memory requirements for DiT training by extending the analytical memory formulation described in \citep{korthikanti2023reducing}. During Aquarius training, xPU memory consumption for model states (parameters, gradients, optimizer states) reaches approximately 300GB, while intermediate activation tensors would scale to around 2TB depending on input video sequence length. Both model states and activation exceed the memory capacity of individual devices. The significant memory gap between model states and activation also serves as a design guideline for navigating trade-offs between parallelism strategies and memory optimization techniques.

\subsection{Parallel Strategy}

\textbf{Innermost TP-SP.} Among 8 ranks within a node, we adopt Tensor Parallelism(TP) \citep{shoeybi2019megatron} and Sequence Parallelism(SP) \citep{korthikanti2023reducing} to partition the model states and activations. In each Transformer block, we employ a dedicated MLP (a Linear layer followed by a SiLU) to incorporate conditioning embedding for Adaptive Layer Normalization (AdaLN), aiming to enhance the model's capacity to capture conditional information. However, this design incurs substantial memory overhead: these dedicated Linear collectively increase the model's total parameter count by over 3B and introduce redundant model states exceeding 20GB per rank. To address this issue, we partition these Linear layers in a column parallel fashion with their computation and parameters parallelized across devices. Specifically, during the forward pass, all-gather reconstructs the complete Linear output, while reduce-scatter synchronizes gradients across the sequence-parallel group before the Linear in the backward pass.

\textbf{Mid-Optional CP.}  While TP-SP supports training with moderately long contexts, Context Parallelism (CP) \citep{jacobs2023deepspeed} is optionally enabled only for ultra-long sequences (exceeds 200k tokens). For non-attention computations, each layer processes 1/CP of the input sequence. Before and after attention computations, all-to-all communication is applied to transpose the sequence, enabling each rank to process attention with full context length but only 1/CP of the attention heads. Since CP incurs lower communication overhead than TP-SP, we configure TP-SP as inner parallelism and CP as outer parallelism.

\textbf{Outermost ZeRO-DP.} We employ DP as the outermost parallelization to enhance throughput. Since the exponential-moving-average (EMA) parameters and moments in AdamW contribute substantially to memory footprint, we adopt ZeRO \citep{rajbhandari2020zero} optimization to partition redundant optimizer states across data-parallel groups. Compared to optimizer states, memory consumption for model parameters and gradients proves relatively minor. To minimize cross-node communication overhead, we maintain full replicas of parameters and gradients during forward/backward passes(ZeRO1 fashion). This reduces cross-node all-gather and reduces-scatter operations to once per gradient accumulation step, amortizing communication overhead.

\textbf{Communication-Computation Overlap.} For TP-SP, the matmul computations is fused with all-gather/reduce-scatter communication as a fusion operation. Within the fused operation, large computation and communication tasks are divided into finer-grained subtasks. These subtasks are then scheduled in a pipelined manner, enabling mutual overlap between communication and computation to minimize latency. For DP, the parameter all-gather is scheduled to overlap with the first forward micro-step, while the gradient reduce-scatter overlaps with the final backward micro-step.

\textbf{Explicit Tensor Parallel Synchronization for patchify.} In Tensor-Parallel (TP) training, non-partitioned parameters are prone to parameter drift across ranks within the TP group. Specifically, parameters in non-partitioned layers are initialized identically across all ranks but might gradually diverge after multiple training iterations. In our experiments, this phenomenon occurred in DiT's patchify layers, resulting in abnormal video outputs.

\begin{figure}[htbp]
  \centering
  \includegraphics[width=0.6\textwidth]{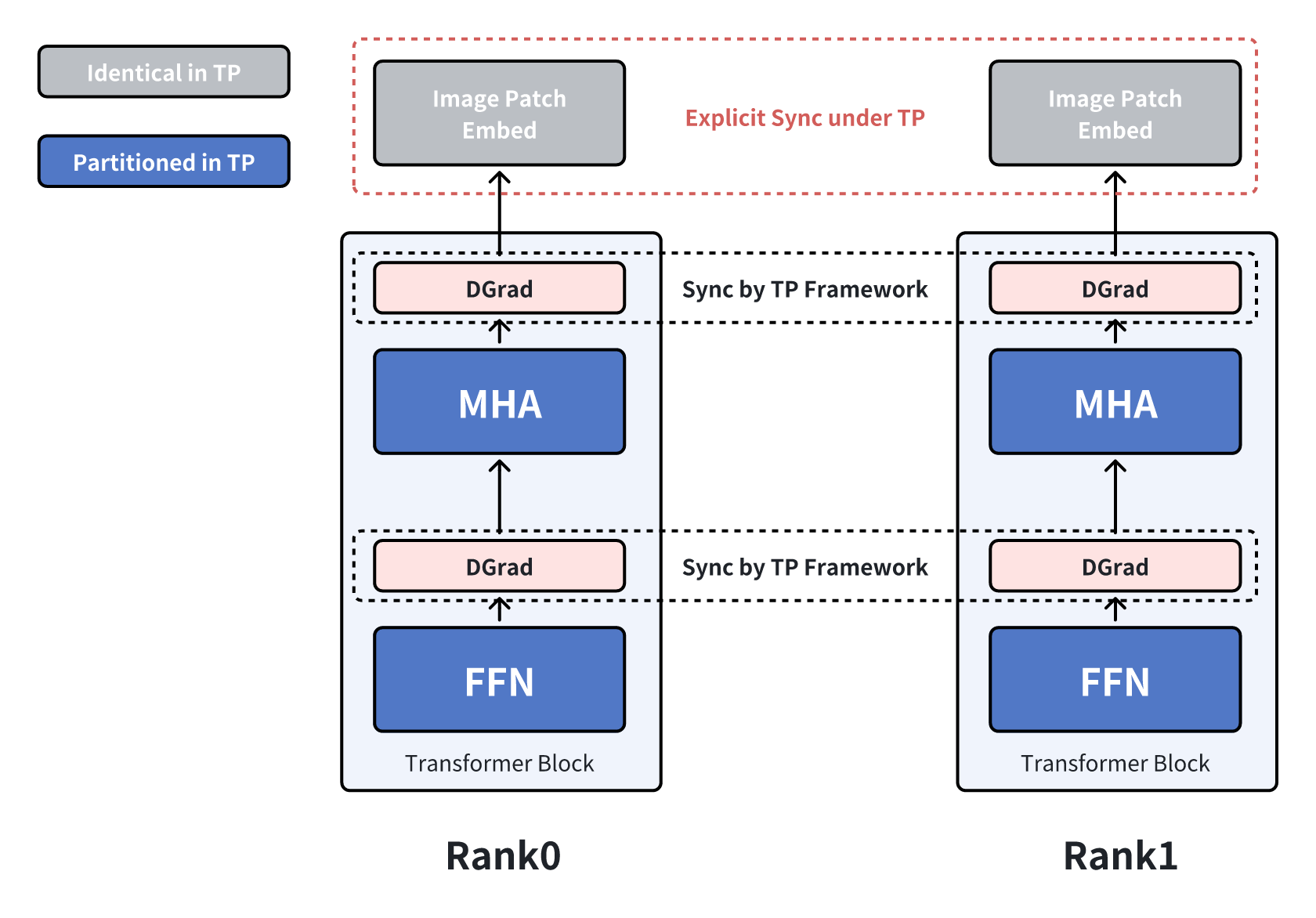}
  \caption{Explicit gradient synchronization across tensor-parallel group: preventing parameter drift in image patchify layers.}
  \label{fig:tp_sync}
\end{figure}

The root cause stems from the synchronization mechanism of tensor parallelism. By default, only gradients of intermediate activations are synchronized, whereas gradients of non-partitioned parameters are neglected. Within the TP group, numerically unstable operations (e.g., atomic floating-point operations) generate slightly divergent gradients under identical inputs. These gradient discrepancies accumulate over training steps, ultimately leading to parameter drift. Parameters within the Transformer structure are either partitioned via tensor parallelism or synchronized via sequence parallelism, thus avoiding this issue. However, external modules (e.g., patchify layers) lack such synchronization mechanisms, becoming primary sources of parameter divergence.

To resolve this issue, we introduced explicit gradient synchronization for these layers within the tensor-parallel group during training, ensuring consistent parameter updates across all devices. This modification effectively eliminated parameter drift in those layers and resolved the abnormalities in the output video.

\subsection{Memory Optimization}

To maximize memory utilization efficiency, fine-grained control over memory optimization techniques—including activation tensor lifecycle management, recomputation, and offloading—proves essential.  These optimizations require stage-specific adaptation to align with the distinct memory consumption patterns across training stages characterized by sequence-length variations.

\noindent

\textbf{Activation Tensor Lifecycle Control.} Long-context training encounters moderate memory bottlenecks due to delayed deallocation of intermediate activations, particularly in two scenarios: \textbf{(1) Shared storage retention}—when multiple tensors (e.g., from split/fused kernels) reference the same underlying storage, earlier tensors' memory remains occupied until the last referencing tensor is released; \textbf{(2) Merged tensor redundancy}—small tensors concatenated  (e.g., via concat/stack/all-gather) retain their original memory allocations despite being obsolete post-merger. By identifying these patterns in computational graphs, we achieved fine-grained memory lifecycle control to trigger prompt deallocation, reducing peak memory usage.

\begin{figure}[htbp]
  \centering
  \includegraphics[width=0.6\textwidth]{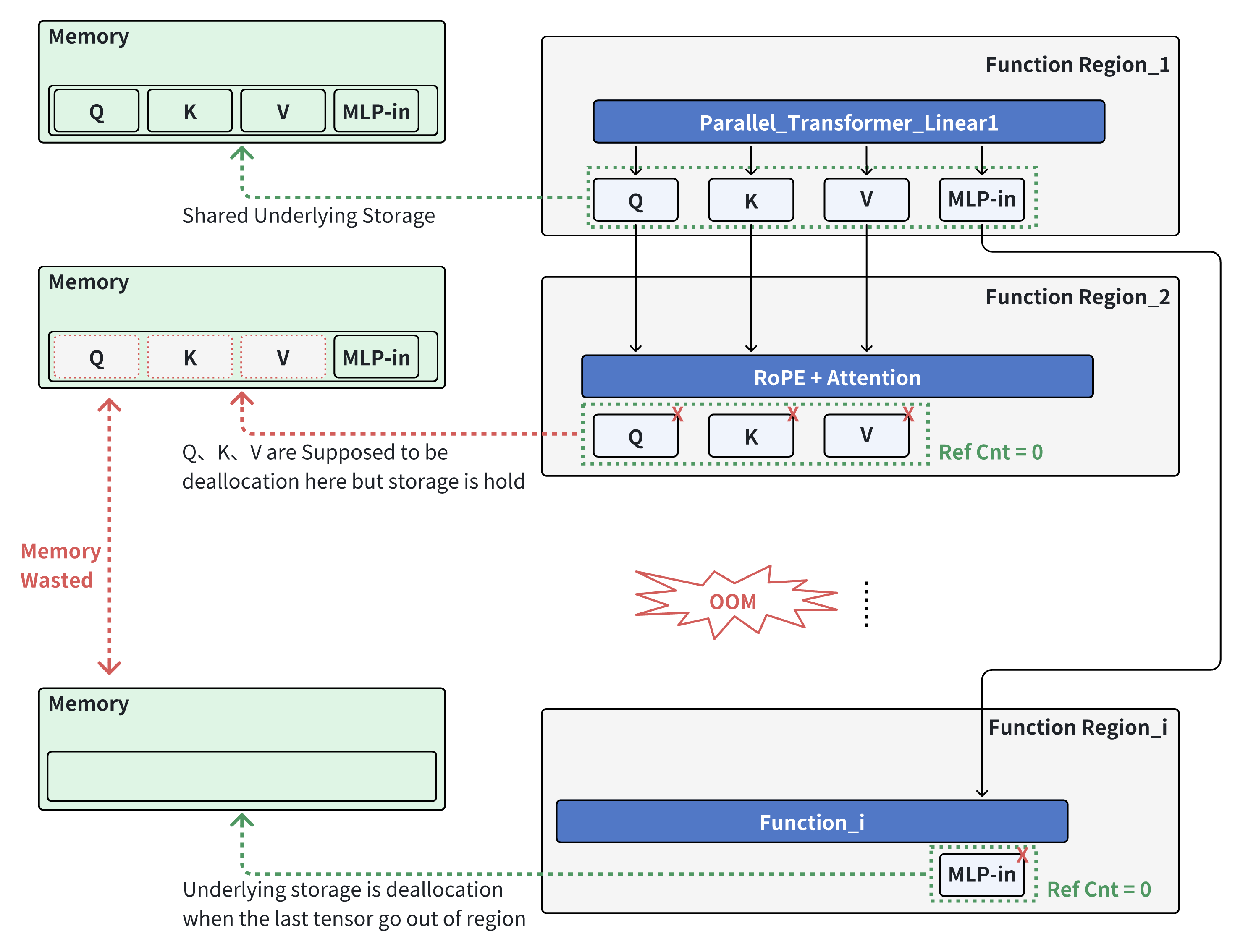}
  \caption{Shared Storage Retention: QKV memory deallocation delayed until MLP-in Release.}
  \label{fig:gc}
\end{figure}


\textbf{Selective Recomputation.}
We adopt the Selective Activation Recomputation methodology in \citep{korthikanti2023reducing} to reduce activation memory overhead.  Diverging from the original approach, we found that recomputing only the core attention part of the Transformer block does not deliver the most efficient performance in long-context scenarios:

\begin{itemize}
    \item The built-in recomputation in Flash Attention's \citep{dao2205fast} backward pass eliminates materialization of attention matrices that exhibit quadratic scaling with sequence length, thereby reducing its memory footprint.
    \item In long-context scenarios \((S \gg H)\), the \( O(S^2H)\) computational cost of Attention operations dwarfs Linear layers' \(O(SH²)\) and element-wise operators' \(O(SH)\).
\end{itemize}

In our experiments, the Transformer block is divided into fine-grained chunks. We evaluate the recomputation memory-latency ratio (memory saved per latency cost) for each chunk and select those with the highest ratios across the computational graph for recomputation. Specifically, based on input tensor size, we calculate the memory consumption of activations retained and profile the actual training latency for operations in forward. An example of memory-latency ratios in Aquarius is shown in Table below. As demonstrated, recomputing operators such as GeLU incur significantly lower computational costs than recomputing layers such as Flash Attention, while achieving equivalent xPU memory savings. An empirical guideline is that recomputing IO-bound operators typically provides a better memory-latency ratio than recomputing compute-bound operators.

\begin{table}[htbp]
\centering
\begin{tabular}{lcrr}
\toprule
Operations & \multicolumn{1}{c}{Activations Retained (MB)} & Forward Latency (ms) & Ratio \\
\midrule
Flash Attention            & $(2BSH+64BAS) / TP = 106$     & 127.5  &   0.8 \\
Out\_Linear + ReduceScatter & $2BSH / TP = 84$              &  13.4  &   6.3 \\
FFN\_Linear2 + ReduceScatter& $2BSH / TP = 84$              &   8.9  &   9.4 \\
AllGather + FFN\_Linear1    & $8BSH / TP = 337$             &   8.6  &  39.2 \\
AllGather + QKV\_Linear     & $6BSH / TP = 252$             &   7.7  &  32.7 \\
Fused QKNorm                & $4BSH / TP = 168$             &   1.9  &  88.4 \\
Gate                        & $2BSH / TP = 84$              &   0.36 & 233.3 \\
LayerNorm+Scale/Shift       & $4BSH / TP = 168$             &   0.58 & 289.6 \\
GeLU                        & $8BSH / TP = 337$             &   0.64 & 526.6 \\
\bottomrule
\end{tabular}
\caption{Recomputation memory-latency ratios in 125-frame 720\texttimes1280 video (seqlen=115k tokens).} 
\label{tab:recompute_metrics}
\end{table}

Building upon memory-latency analysis, we implement a heuristic greedy algorithm that selects a set of operations with minimal cost for recomputation while guaranteeing no memory overflow occurs during training. By implementing custom backward passes via \textit{torch.autograd.Function}, we achieve fine-grained recomputation control at operation level with negligible computational overhead.

During different training stages, significant variations in video duration and resolution lead to substantial differences in memory consumption across stages. For each training stage, we recalculate the recomputation cost-benefit ratios based on current-stage input sizes and reselect the optimal operator set for recomputation. This ensures minimal recomputation overhead at each stage while maintaining compliance with stage-specific memory constraints.

\textbf{Offloading.} We further alleviate device memory bottlenecks by adopting offloading strategies \citep{rhu2016vdnn,zhu2024zerof} that leverage host memory for storing non-critical tensors. 

In certain downstream training tasks with smaller Data Parallelism (DP) scale, the xPU memory footprint of optimizer states emerges as a critical bottleneck. To mitigate this, we offload optimizer states to host memory during forward and backward computations, transferring them back to device memory exclusively during the optimizer update phase. This strategy effectively reduces peak device memory consumption during forward and backward phases while preserving computational integrity, as optimizer states reside in device memory only when strictly required. Under large gradient accumulation steps, the frequency of optimizer state data transfers is reduced. Furthermore, we can overlap Device-to-Host (D2H) transfers with computations in the first forward micro-step, and overlap Host-to-Device (H2D) transfers with computations in the last backward micro-step, effectively reducing the overhead of optimizer state offloading.

Activation offloading is adopted for long-context scenarios where intermediate activation tensors exceeding predefined thresholds (those exempt from recomputation) are offloaded to host memory. During the forward phase, D2H transfers of the current Transformer block's activations overlap with the next block's forward computation. In the backward phase, H2D transfers overlap the previous block's gradient computation.  Concurrent accesses from all devices within a single NUMA domain strain host DDR write bandwidth, creating a bottleneck in PCIe throughput that constrains the applicability of offloading strategy in our experiments. Future work will explore device-interleaved scheduling to alleviate cross-device bandwidth contention.

\textbf{Strategies Balancing.} Achieving optimal training performance requires balancing three memory optimization techquies—parallelism (trading inter-device communication for memory), recomputation (trading computation for memory), and offloading (trading PCIe communication for device memory)—which is a non-trivial challenge due to the vast combinatorial search space. Based on our empirical observations, we follow these guidelines:

\begin{itemize}
    \item Offloading is employed when data transfers can be overlapped with computation. Appreciable overlap potential exists in long-context scenarios, whereas shorter context cases exhibit severely constrained overlap capacity.
    \item Fine-grained coordination between recomputation and offloading is essential: offloading eliminates recomputation for expensive compute-intensive operators (e.g., attention), whereas recomputation is strategically applied to activation tensors with lower recomputation costs.
    \item Using the two aforementioned techniques to constrain context parallelism strategies to a minimally viable degree, as cross-node all-to-all communication constitutes the major bottleneck in our experiments. 
\end{itemize}

The strategy configurations would vary depending on training and hardware specifications. In future work, we plan to automate the exploration of strategy space via modeling and profiling, leveraging domain-specific heuristics(e.g., hardware bandwidth constraints) to pruning the search space. 

\subsection{Training Efficiency and Stability in Cluster}

In large-scale training environments, anomalies in software and hardware layers within clusters can severely degrade training efficiency. By leveraging the robust training techniques—including self-check diagnostics, monitoring systems, and fast recovery mechanisms—integrated in cluster platform \citep{jiang2024megascale}, our training achieves high training stability and efficiency in large-scale clusters.

In production cluster training scenarios, fragmented task scheduling often degrades communication performance. Specifically, we observed elevated Priority Flow Control (PFC) metrics and persistent network congestion during large-scale training, leading to cross-node communication bandwidth falling below the hardware's theoretical capacity.

By adopting a topology-aware scheduling strategy and reordering node IP sequences, we enhanced task topology affinity, thereby reducing cross-TOR (Top-of-Rack) traffic and improving cross-node communication throughput during large-scale data-parallel training.

Potential stragglers are proactively eliminated through custom diagnostics prior to formal training. Through low-cost, coarse-grained metrics collection across training phases (forward, backward, updates), we pinpoint and resolve performance fluctuations caused by irregular memory defragmentation within the framework, ensuring efficient large-scale cluster training.

\section{Inference Strategy}
\label{section6}
Model Inference acceleration can be achieved through two approaches: Training-based and Training-free. Training-based methods primarily involve classifier-free guidance \citep{ho2022classifierfreediffusionguidance} (CFG) distillation, which integrates the inference effects of positive and negative prompts into a single step, or leverage distillation techniques \citep{ren2024hypersdtrajectorysegmentedconsistency} to reduce number of function evaluations (NFE). However, both approaches demand meticulous model architecture design and training efforts. To more flexibly and rapidly address on-the-fly deployment scenarios and to decouple the training and inference processes, we have adopted the Training-free acceleration methods for inference.

\textbf{Diffusion Cache.} Inspired by Training-free methods such as \citep{chen2024deltadittrainingfreeaccelerationmethod, zhao2025realtimevideogenerationpyramid}, we observed analogous phenomena in the Aquarius model:
\begin{itemize}
    \item \textbf{Feature similarity in DiT layers} Front DiT blocks focus on generating image contours, while rear DiT blocks specialize in refining details.  During different stages of the sampling process, the feature deviations (outputs) from rear DiT blocks exhibit significant similarity across sampling steps.
\end{itemize}

\begin{itemize}
    \item \textbf{Attention similarity} The attention outputs of the same DiT block show significant similarity across different sampling steps.
\end{itemize}

Specifically, we select certain steps for caching by comparing the quality of generated videos after evaluating the results from different sampling step intervals and caching strategies. We achieve this by randomly sampling prompts from the validation set and enumerating the DiT layers, caching sampling step intervals, and sampling steps.

\begin{itemize}
    \item \textbf{DiT layer output caching} For the backend DiT layers within the sampling steps, we execute the output feature offsets of multiple DiT layers every few steps and reuse the cached results in subsequent steps, overlaying offsets on the output features of the frontend DiT layers.
\end{itemize}

\begin{itemize}
    \item \textbf{Attention cache} For the attention components, we perform the forward pass of the cached attention module output every few steps and reuse the cached results in the subsequent steps.
\end{itemize}

In the Aquarius model, we found under the premise of ensuring the generation quality on the validation set and with the same acceleration ratio, DiT caching exhibits more stable performance than attention caching. DiT can enhance inference performance by 1.67 times and has good parallelism in multi-xPU settings. It is important to note that, for more stable generation results, no caching strategies are employed in the first ten denoising steps.

\textbf{VAE parallel.} In the VAE decoder stage, we assign different latent tiles to different xPUs. After parallel computation, we obtain the final values by linearly summing the overlapping parts, achieving a linear speedup ratio.

\textbf{DiT parallel.} We follow the same parallelization strategy as during training. Within each node, we employ TP-SP (Tensor Parallelism - Sequence Parallelism) to partition model parameters and intermediate activations across the 8 xPUs, thereby reducing the single-step latency of inference. Between nodes, we utilize multi-xPU data parallelism to enhance overall throughput.

\begin{figure}[htbp]
    \centering
    \begin{subfigure}[t]{0.48\textwidth}
        \centering
        \includegraphics[width=\linewidth]{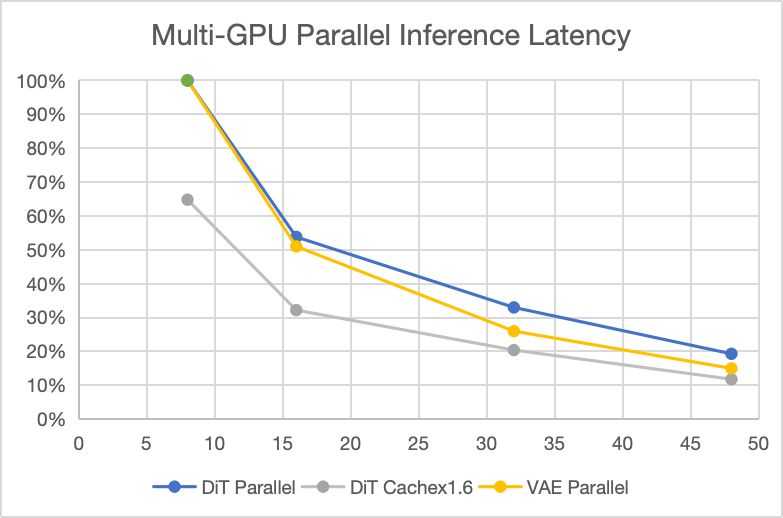}
        \caption{multi-xpu-infer-latency}
        \label{fig:side:a}
    \end{subfigure}
    \hfill 
    \begin{subfigure}[t]{0.48\textwidth}
        \centering
        \includegraphics[width=\linewidth]{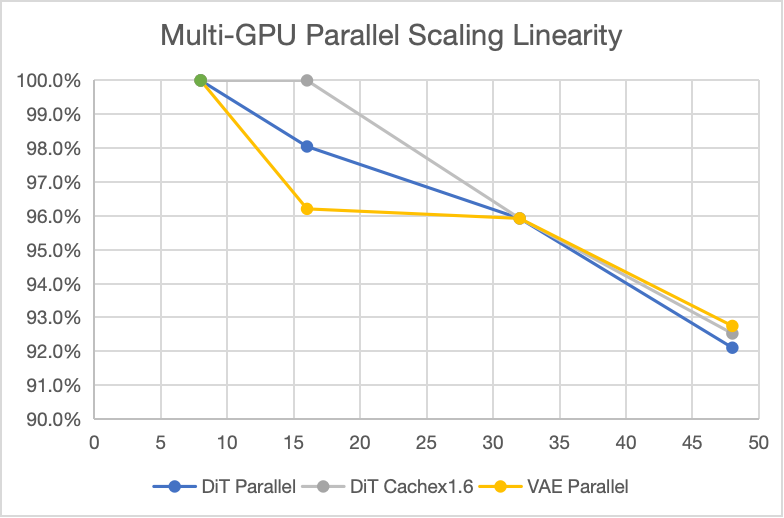}
        \caption{Multi-xPU Parallel Scaling Linearity}
        \label{fig:side:b}
    \end{subfigure}
    \caption{Multi-xPU Parallel Inference Scaling Linearity}
    \label{fig:side}
\end{figure}

\section{Applications}
\label{section7}
In commercial scenarios, users require a substantial supply of high-quality video clips to enhance the richness, production efficiency, and conversion efficiency of advertising creatives. Video generation technology can significantly reduce the cost of creative production, improve production efficiency, and increase the diversity of the overall advertising creative ecosystem due to the inherent randomness of video generation. This approach is user-friendly and achieves a win-win situation for the platform, advertisers, and users.

However, the commercialization and real-world deployment of video generation technology face unique challenges.

\textbf{Scene Incompatibility:} General  video generation models prioritize aesthetic composition and textural fidelity, whereas advertising/marketing content demands scenario-driven narratives that resonate with real-world consumer contexts.  Therefore, they emphasize more on down-to-earth, lifelike scenes.  There is a gap in the general model's ability to fit well with real-life scenarios and meet the requirements of advertising videos.

\textbf{Generated Avatar Unrealistic:} Insufficient Realism in Generation: Current general models fall short of the commercial scene's demands in terms of character authenticity, motion realism, and natural lighting.   For example, the character images tend to be more Westernized, have a strong AI feel, feature monotonous movements and expressions, and the lighting and shadows do not match the commercial scenarios.

Aquarius has been tailored to address the specific challenges within advertising scenarios and has engaged in the exploration and deployment of generative techniques across various downstream applications.

\subsection{Digital Human Generation}
Aquarius is specifically optimized for commercial character generation scenarios, offering enhanced prompt-video alignment capabilities. Users can leverage carefully engineered prompts to precisely control attributes such as appearance, attire, and contextual interactions, thereby customizing and generating digital human video assets tailored for commercial applications like branded storytelling, product demos, or marketing campaigns. 

\begin{figure}[htbp]
  \centering
  \includegraphics[width=\textwidth]{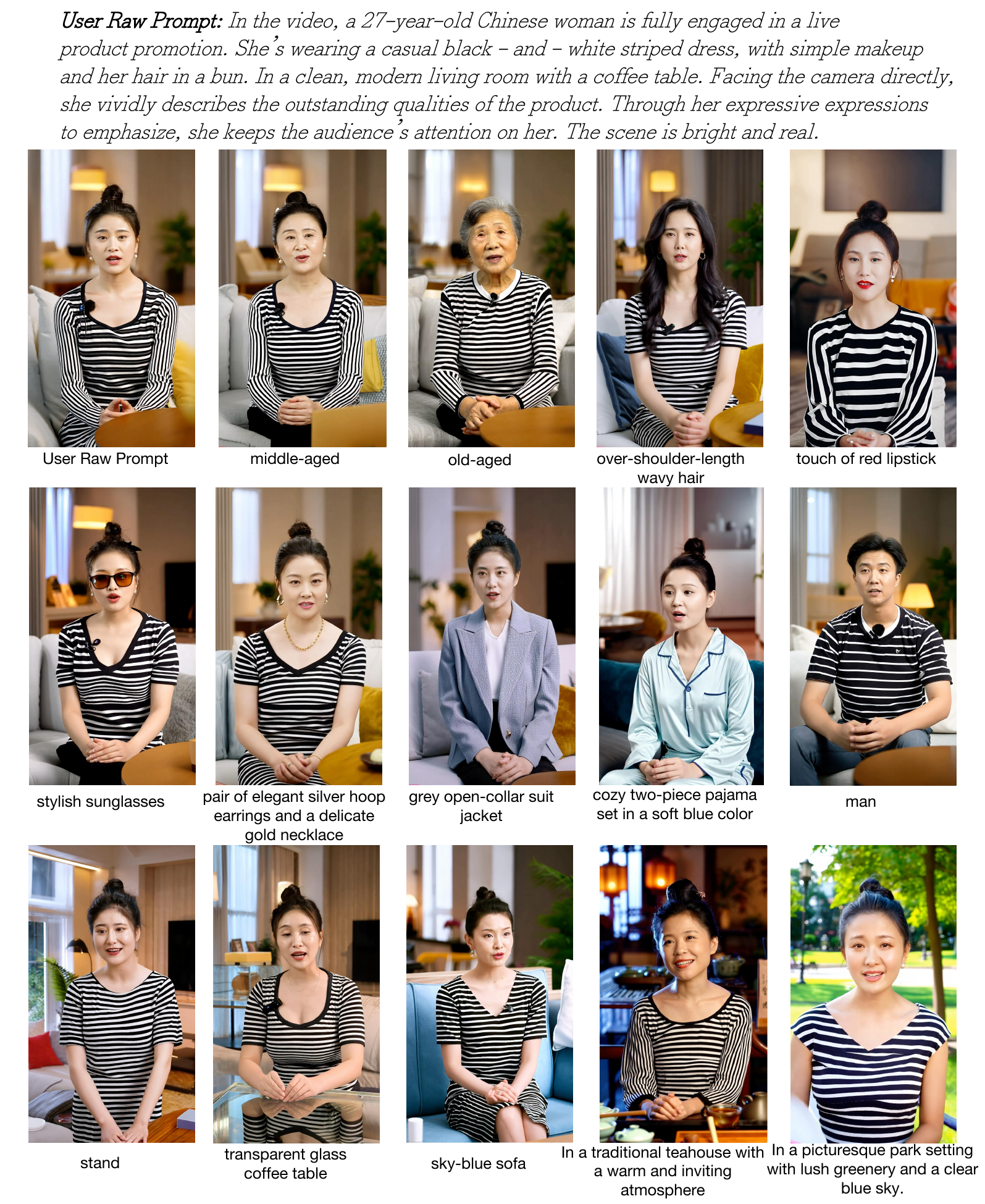}
  \caption{The 2B single-DiT model generates a 5-second video at 1080P resolution with 24 frames per second (fps): The first video in the top-left corner represents the original video generated from the user's raw prompt. The subsequent videos verify prompt-video alignment by modifying only a single attribute of the user's raw prompt. This reflects the model's strong instruction-following ability and fine-grained video generation capabilities.
}
  \label{fig:attribute}
\end{figure}

\textbf{Fine-grained Control Capability.} As shown in Figure \ref{fig:attribute}. Leveraging the video generative model's robust text comprehension and instruction alignment capabilities, users can customize the video character's attributes such as gender, age, hairstyle, accessories, clothing, and makeup through prompts, generating digital human videos that better fit their specific commercial scenarios.

\begin{figure}[htbp]
  \centering
  \includegraphics[width=\textwidth]{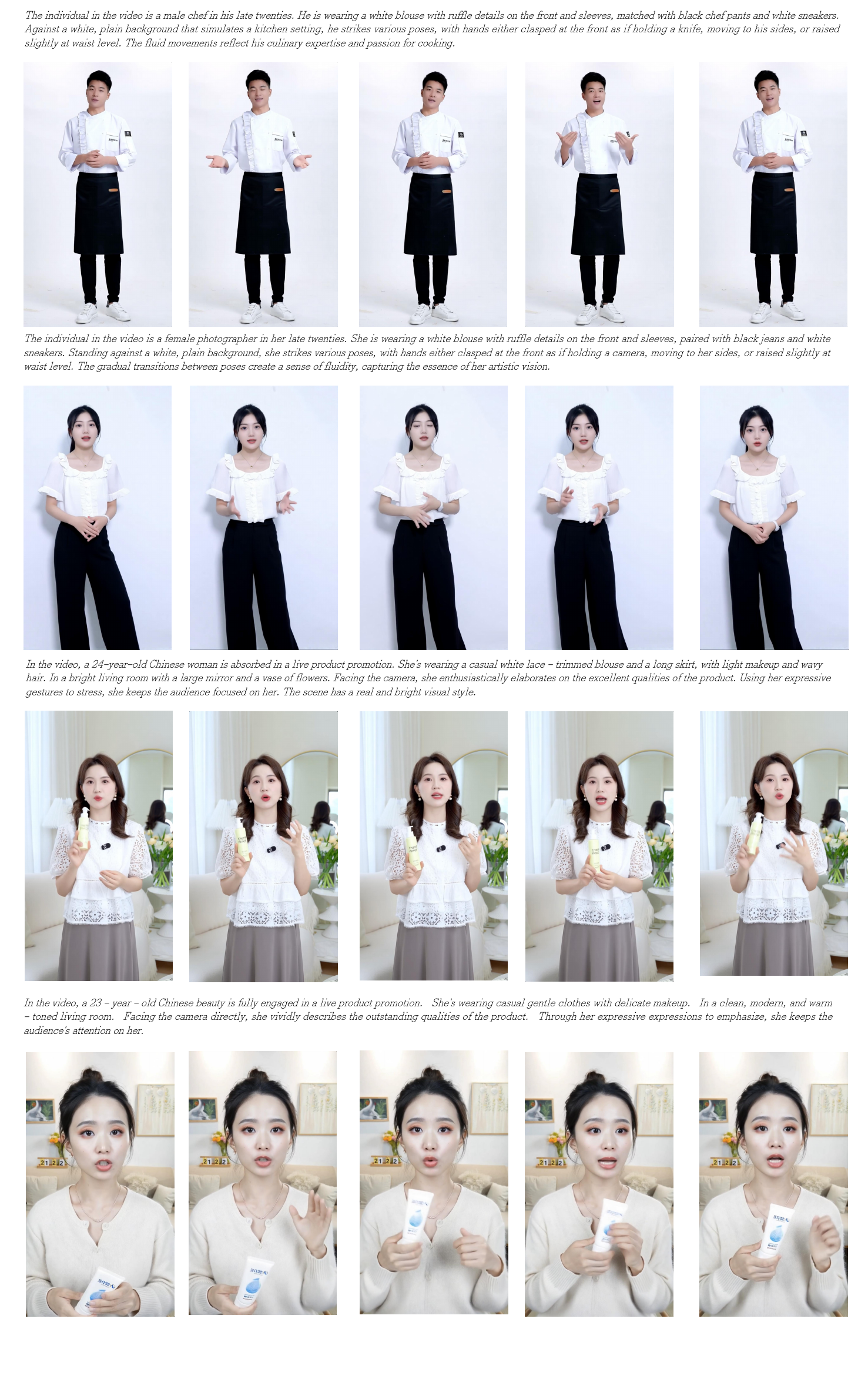}
  \caption{The 13.4B Multimodal-DiT model generates a 5-second video at 720P resolution with 24 frames per second (fps).
}
  \label{fig:digitalman_2}
\end{figure}

\begin{figure}[htbp]
  \centering
  \includegraphics[width=\textwidth]{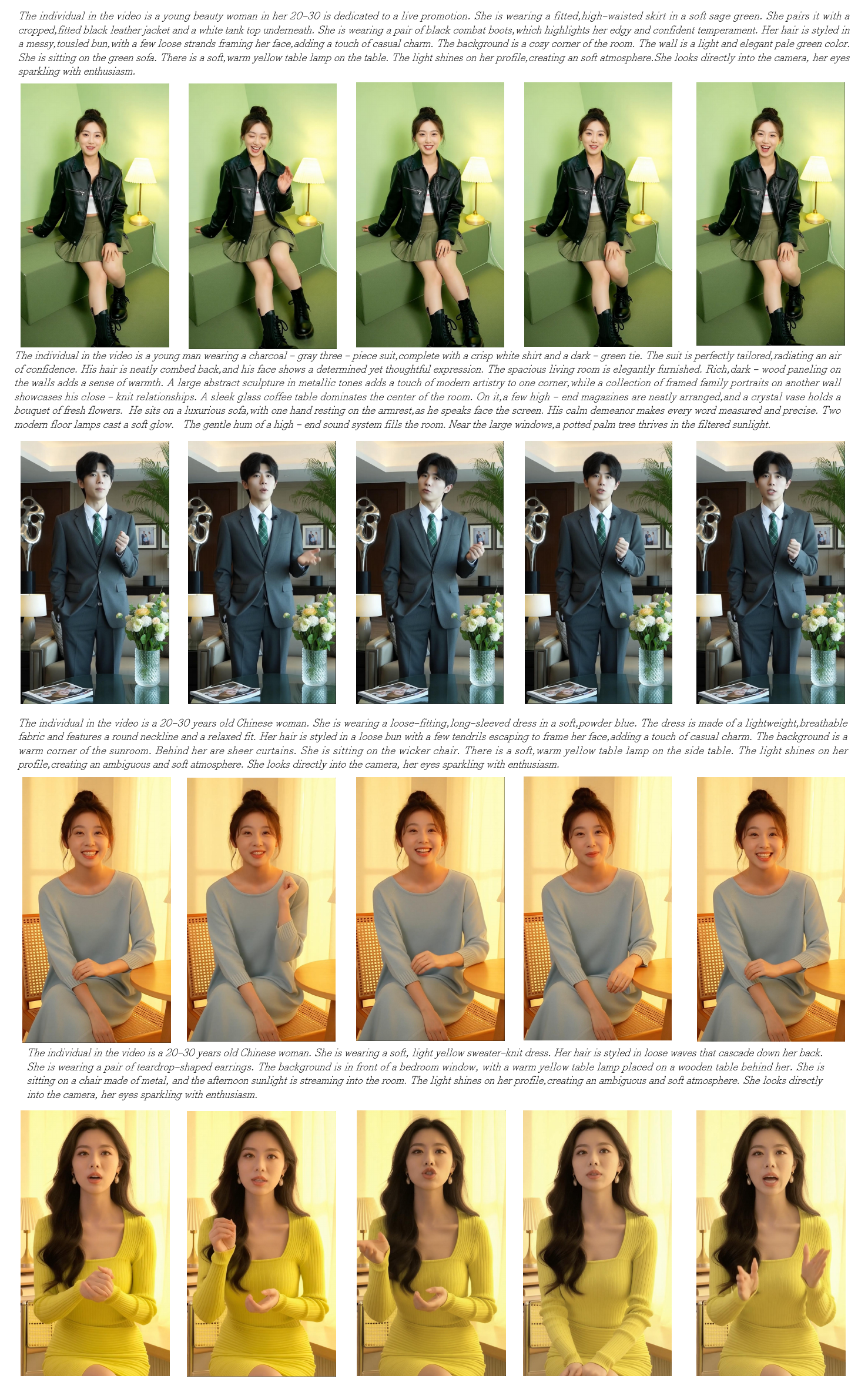}
  \caption{The 13.4B Multimodal-DiT model extended generates a 10-second video at 720P resolution with 24 frames per second (fps).
}
  \label{fig:digitalman_1}
\end{figure}

\textbf{Realistic and Natural Motion.} As shown in Figure \ref{fig:digitalman_2}, \ref{fig:digitalman_1}. With just a short prompt, the video generation also can automatically generate background details, producing professional digital human videos with whiteboard backgrounds, as well as digital humans for product promotion. The video quality are realistic and the movements are natural, which perfectly fits the needs of commercial scenarios in downstream applications.

\subsection{Video Inpainting}
Video inpainting aims to fill in plausible pixels in the missing video area. It has a wide range of applications in the field of video editing, such as video decaptioning, object removal and video completion. Video inpainting is challenging because it is supposed to generate video contents which are visually realistic as well as spatial-temporal consistent with surroundings. In addition to that, one needs to inpaint pixels for all frames in the video, which is usually time-consuming and hinders practical application.

To address above concerns, we present an efficient video inpainting model based on Diffusion Transformer. Unlike most video editing methods recently, which use a large pretrained model and finetune from it for downstream applications, our model adopts a self-designed small transformer-based network which is trained from scratch. To deal with long videos, we employ MultiDiffusion \citep{bartal2023multidiffusionfusingdiffusionpaths} for the temporal consistency of transition frames, making it convenient to apply to video decaptioning and video completion tasks. Experiments show that our method can produce competitive results with an acceptable time cost, compared to existing video inpainting algorithms.

\begin{figure}[h]
    \centering
    \includegraphics[width=1.0\textwidth]{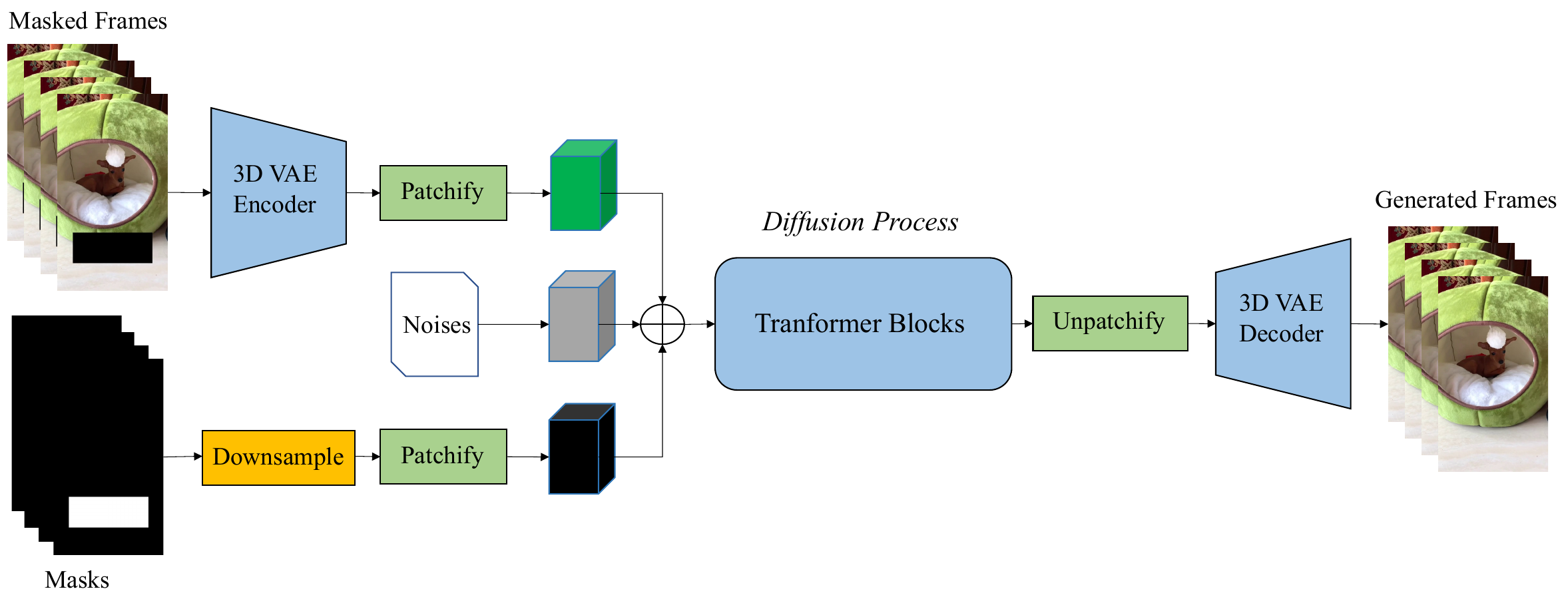}
    \caption{Pipeline of video inpainting. We patchify video latents, masks along with random noises and add them together as a sequence of tokens. After the diffusion process conducted through several transformer blocks, we can decode tokens into video frames as our final results.}
    \label{fig:pipeline}
\end{figure}

\textbf{Model Design.}
Given a masked video sequence $Y \in \mathbb{R}^{H\times W\times N \times 3}$ with $N$ frames, along with corresponding masks $M \in \mathbb{R}^{H\times W\times N \times 1}$, video inpainting aims to generate plausible visual contents in the masked area which should be consistent and coherent with surrounding pixels. At first, we use 3D VAE to encode $Y$ into video latents $y$ and downsample its time-space dimensions, as $y \in \mathbb{R}^{h\times w\times n \times c}$. Then latents $y$, downsampled masks $m$ and noises with the same size are all fed into a transformer-based network. After several denoising steps, video latents output by Diffusion Transformers are decoded into the final results $X \in \mathbb{R}^{H\times W\times N \times 3}$. For varying lengths of video sequences, we utilize MultiDiffusion in the temporal axis for the consistency of transition frames between video clips. Figure \ref{fig:pipeline} illustrates the inference pipeline of our method.

\textbf{3D VAE.} We use a 3D VAE to encode masked frames $Y$ into the latent space for video compression. Given masked video frames $Y \in \mathbb{R}^{H\times W\times N\times 3}$, the encoder compresses them into low-dimensional latents $y \in \mathbb{R}^{h\times w\times n\times c}$, where $h=H/8, w=W/8, n=(N-1)/4+1$ and $c=8$. Corresponding masks $M \in \mathbb{R}^{H\times W\times N\times 1}$ are also downsampled to $m$ with the same size $h \times w \times n$. We find in experiments that missing temporal information of masks may lead to flickers in inpainting results. Therefore, we recover the reduced temporal dimension of masks through the channel dimension for completeness, as $m \in \mathbb{R}^{h\times w\times n\times 4}$. In the decoding stage, the denoised latents are decoded into the original resolution as inpainting frames $X \in \mathbb{R}^{H\times W\times N\times 3}$.

\textbf{Diffusion Transformer.} For masked video latents $y$, downsampled masks $m$ and random noises, we first patchify them of the same $2 \times 2 \times 1$ spatial-temporal size and project all 3D patches into the same embedding dimension. The three embeddings are then flattened into 1D sequences and added together as input tokens for transformer blocks. Therefore, the total length of the input tokens is $w \times h \times n/4$.

We adopt a pre-norm transformer block structure primarily comprising a multi-head self-attention and a feedforward network. Our transformer block excludes the cross-attention as there is no text condition. We regress two sets of scale and shift parameters from timesteps through \textit{adaLN-Zero block}, and then inject them into the self-attention and the FFN separately. After the final transformer block, we project each token into a $2 \times 2 \times 2c$ tensor through a linear layer and unpatchify the 1D sequence back into the original 3D size. 


\textbf{Temporal MultiDiffusion.} To address longer videos with more frames than training, we apply MultiDiffusion to the temporal axis, making it effective for our model to inpaint videos of arbitrary length with temporal consistency. Since the denoising process is performed in the latent space, we consider the video latents $x$ for convenience. 

Given a video with $N'$ frames where $N' > N$, the length of its encoded latents $x'$ is $n'=(N'-1)/4+1$. We segment the latents $x'$ into several overlapping clips by a sliding window with a length of $n$ and a stride of $s$. This process partitions $x'$ into latent clips $\{x^k\}^r_{k=1}$, where $r=\lceil (n'-n)/s \rceil +1$ is the total number of clips. The denoising step is performed on each latent clip and we denote the $k$-th clip as $x^k_t$ at the timestep $t$.

For the $i$-th latent of the temporal index, denoted as $x'[i]$, we can find the set of clips $\mathcal{S}(i)=\{x^k|x'[i]\in x^k\}$ that contain this latent. For each clip $x^k$ in $\mathcal{S}(i)$, we denote the corresponding latent as $x^k[j]$ which is mapped from $x'[i]$. After the timestep $t$ denoising process performed on the clips, we update the value of $x'_t[i]$ by averaging all the corresponding latents:

\begin{equation}
x'_t[i]=\frac{1}{\|\mathcal{S}(i)\|}\sum_{x^k\in\mathcal{S}(i)}x^k_t[j].
\end{equation}

We update the values of all latents using the above formulation and then map them to the corresponding clips. Subsequently, the next denoising step is performed on the basis of the mapping values. Since overlapping clips share the same latent space, they are able to maintain temporal consistency at transition frames.

\textbf{Training Details.} We employ a two-stage coarse-to-fine strategy to train our inpainting model, since we find that convergence is difficult to achieve by training the model directly on high-resolution videos. At the first stage, we train the DiT model on 240p videos to capture spatial and temporal consistency in a coarse manner. At the second stage, we continue to train the model on 720p videos to enhance fine details for the high quality. The DiT model is trained for 500k iterations at the first stage and 200k iterations at the second stage. The batch size of training is 16 and the length of video frames is 65. We use the AdamW optimizer with a constant learning rate of 1e-5. To simulate masks used in video decaptioning and video completion tasks, during training, we generate stationary and moving masks in a random pattern following ProPainter \citep{zhou2023propainter}.

\begin{figure}
    \centering
    \includegraphics[width=0.7\textwidth]{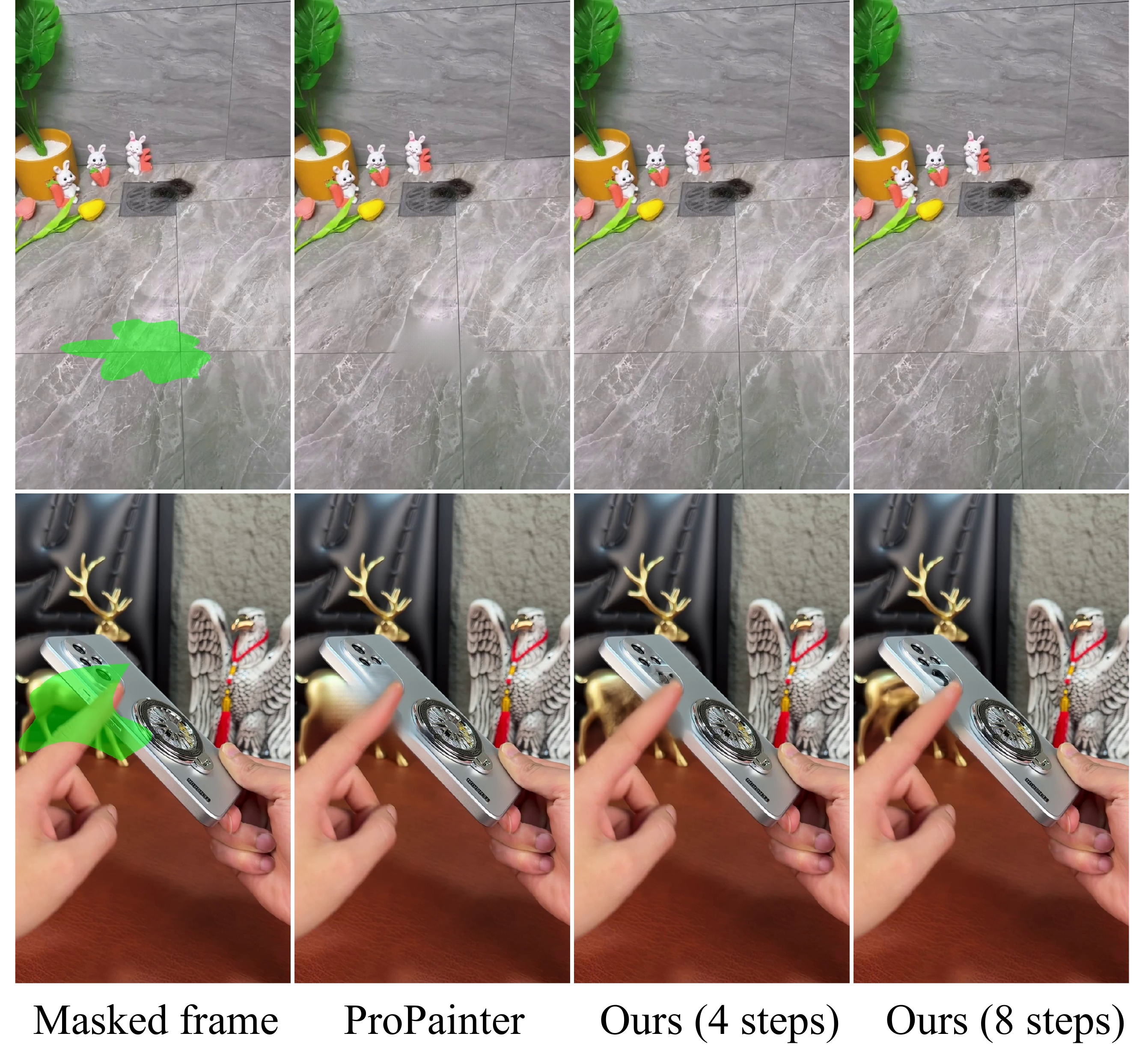}
    \caption{Results of video completion by ProPainter and our method.}
    \label{fig:completion}
\end{figure}

\textbf{Comparison Experiments.} We collect 50 short videos with the resolution of 720p as a test set. We generate masks by the same pattern of training and employ models to complete masked videos to calculate quantitative scores. Video frames are resized to $432 \times 240$ for evaluation. The quantitative scores in Table~\ref{tab:table} show that our method surpasses ProPainter, the state-of-the-art video inpainting algorithm. Due to Flow Matching, our method can address video inpainting even in 4 or 8 inference steps. Figure \ref{fig:completion} visualize some results of video completion in the test set.

\begin{table}
 \caption{Quantitative comparison on video completion.}
  \centering
  \begin{tabular}{llll}
    \toprule
    \cmidrule(r){1-2}
             & PSNR$\uparrow$ & SSIM$\uparrow$ & VFID$\downarrow$ \\
    \midrule
    ProPainter & 34.46  & 0.9834 & 0.069     \\
    Ours(4 steps) & \textbf{34.86} & \textbf{0.9844} & 0.056   \\
    Ours(8 steps) & 34.60 & 0.9843 & \textbf{0.051}  \\
    \bottomrule
  \end{tabular}
  \label{tab:table}
\end{table}

\begin{figure}[h]
    \centering
    \includegraphics[width=1.0\textwidth]{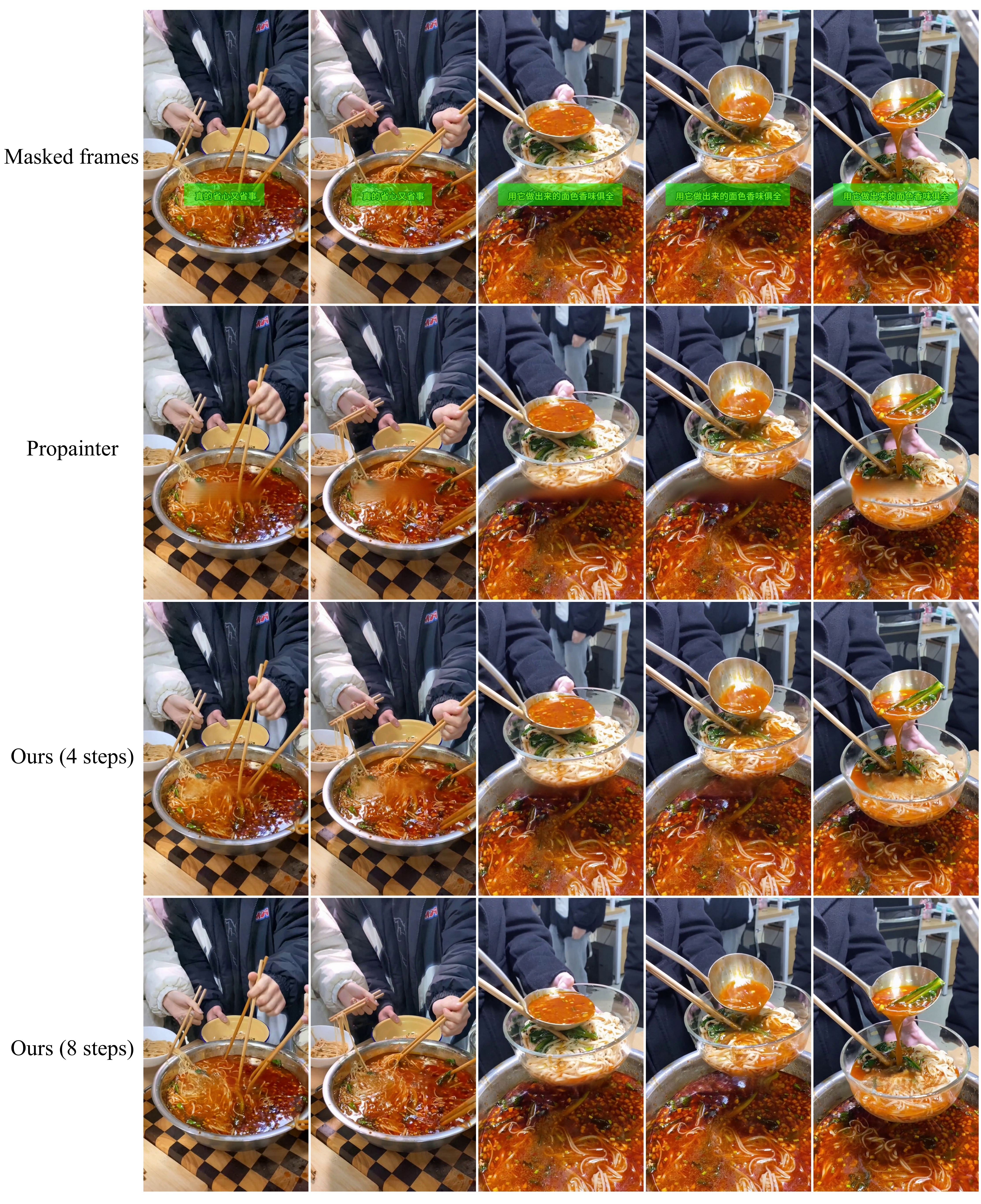}
    \caption{Qualitative comparison of video decaptioning between ProPainter and our method.}
    \label{fig:decaption}
\end{figure}

It is also convenient and efficient to apply our method to the video decaptioning task. Figure \ref{fig:decaption} shows the results of video decaptioning using Propainter and our method. Propainter may cause blurry and artifacts while estimated optical flows are not accurate or there is no corresponding pixel for propagation. In contrast, our method can achieve high-quality results with spatial-temporal consistency and realistic textures.

\subsection{Video personalization}
Video personalization refers to generating videos that maintain consistency with the identity of the reference image. Video personalization has a wide range of applications in advertising scenarios. For example, when creating AI avatars or advertisement videos, it can be used to specify particular character identities, enhancing the appeal of the video characters. 
Based on the pretrained video foundation model, we further integrated an identity preservation module to achieve video personalization capabilities. The following three sections respectively introduce the model design, the process of collecting the training dataset, and the visualization of the model's generated results.

\textbf{Model Design.}
Video personalization is used to generate videos that maintain identity consistency with the reference images. 
Given video frames $Y \in \mathbb{R}^{T \times H\times W\times 3}$ and a reference image $I \in \mathbb{R}^{1\times H\times W \times 3}$, the key model design involves how to extract features from the reference image and inject them into the Diffusion Transformer. Inspired by Concat-ID and WanVideo, we directly use the VAE as the feature extractor, projecting the reference image into the same feature space as the video latent. 
Specifically, the features of reference image $i \in \mathbb{R}^{1\times h\times w \times c}$ are obtained from the VAE encoder. The feature $i$ and the noisy video latent $x \in \mathbb{R}^{t\times h\times w \times c}$ are concatenated in sequence.
$$
x' = {concat}(x,i)
$$
The concatenated features $x' \in \mathbb{R}^{(t+1)\times h\times w \times c}$ are fed into the DiT model for diffusion process. At the output layer of the transformer, only the tokens corresponding to the video latent are decoded to obtain the video.

We also attempted the approach used in HunyuanVideo-i2v and WanVideo-i2v, where the latent representations of the reference image and video are concatenated along the channel dimension. However, due to spatial misalignment, the model training in our experiments struggled to converge, leading to issues such as facial distortions in the generated videos.

\begin{figure}[htbp]
  \centering
  \includegraphics[width=\textwidth]{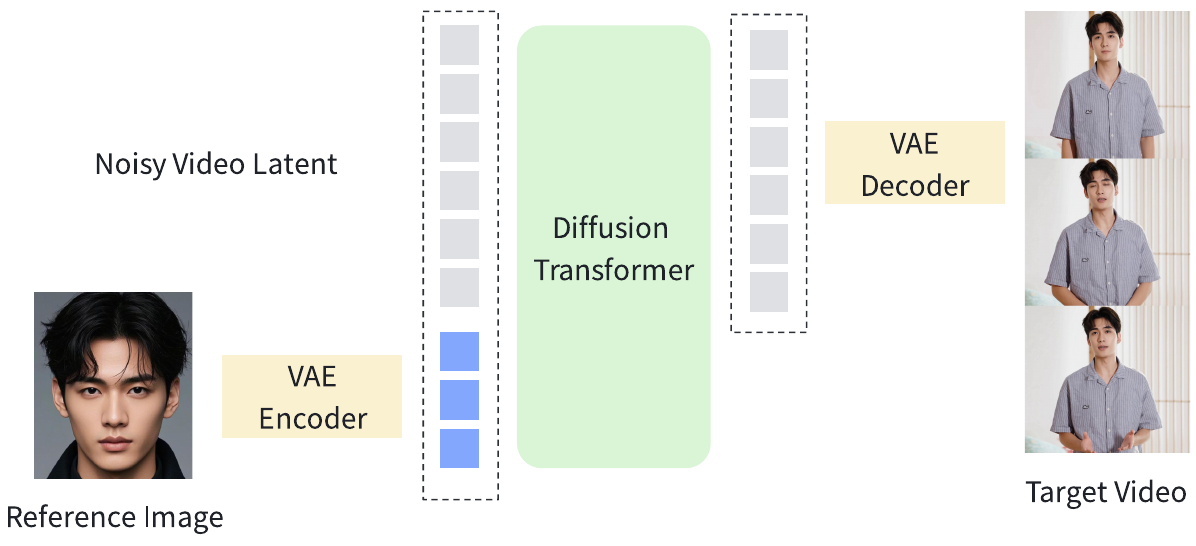}
  \caption{Overview of video personalization model. The reference face image is processed by the VAE encoder to extract features, which are then concatenated with the noisy video latent along the sequence dimension. In the output layer, the tokens corresponding to the video latent are decoded by the VAE to generate the video.}
  \label{fig:face_vis}
\end{figure}

\textbf{Data.}
To construct the training dataset for video personalization, we first segment videos containing people into clips. Then, through face detection, we filter and retain video clips that contain only a single person. Based on the source of the reference images, the training data can be divided into intra-video pairs and cross-video pairs.

\begin{itemize}
    \item \textbf{Intra-video pairs}: In this portion of the data, a random frame is extracted from the video clip, and the face region is detected and cropped as the reference image. To ignore irrelevant areas such as clothing and background, face segmentation is performed, retaining only the face and hair regions.
    \item \textbf{Cross-video pairs}: Models trained solely with intra-video pairs often excessively preserve facial orientation, expressions, and other information from the reference image in the generated videos. Therefore, we selected reference images from different video clips of the same long video to form cross-video pairs. To ensure consistency in person identity, we first calculate the similarity of face embeddings across different video clips and randomly select a frame from the video clip with a similarity greater than 0.7 as the reference image.    
\end{itemize}

In our experiments, we initially trained the model using intra-video pairs, followed by fine-tuning with a mix of intra-video pairs and cross-video pairs.

\textbf{Visualization Results.}
In Figure \ref{fig:face_vis}, we present the results of video personalization. The first column on the left shows the reference images, while the right side displays the generated video results. It can be seen that our model is able to maintain identity consistency while achieving high video quality and good integration with the scene.

\begin{figure}[htbp]
  \centering
  \includegraphics[width=\textwidth]{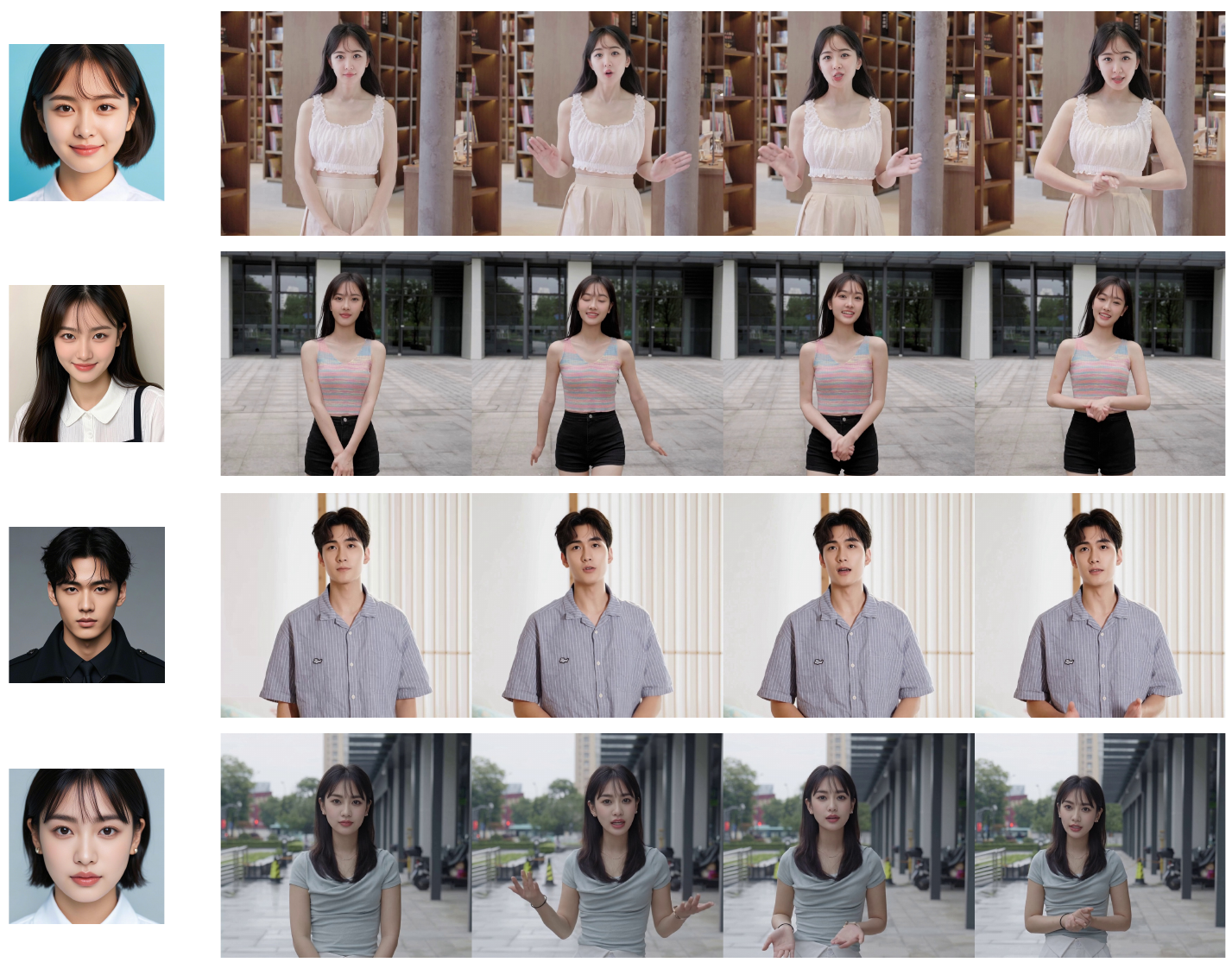}
  \caption{Visualization results of video personalization model.}
  \label{fig:face_vis}
\end{figure}

\section{Conclusions}
\label{section8}
In this report, we propose Aquarius, a family of industry-level video generation models for marketing scenarios designed for thousands-xPU clusters and models with hundreds of billions of parameters. We demystify the black box of the video generation system by presenting its model architecture, training details, and infrastructure construction. We demonstrate various applications of video generation capabilities in commercial scenarios. To further lower the barrier for establishing key data processing workflows, we will open-source Aquarius-Raydata to promote the growth of the open-source community.

\section{Future Works}
\label{section9}
\textbf{RLHF.} Diffusion-based generative models synthesize highly detailed and realistic samples through an iterative denoising process, yet minor deviations in early denoising stages can propagate and amplify across subsequent iterations.  For instance, directional errors occurring during initial denoising phases may ultimately cause the generated imagery to deviate significantly from intended outcomes.  Empirical observations reveal that prolonged supervised fine-tuning (SFT) with high-quality datasets paradoxically induces performance degradation despite initial improvements.  Furthermore, in visual generation tasks, human evaluation metrics consistently demonstrate superior reliability compared to automated scoring systems.  This necessitates implementing reinforcement learning from human feedback (RLHF) post-SFT to align model outputs with anthropocentric quality assessments, thereby bridging the critical gap between algorithmic generation and human perceptual expectations.

\textbf{DiT-MoE.} Designing efficient routing strategies that effectively perceive spatial redundancy and adapt to temporal complexity variations in diffusion tasks remains a challenge in our method when integrating MoE architectures into diffusion transformers. Future work is required to improve expert assignment for better computational resource allocation in diffusion tasks.

\textbf{Unified Understanding and Generation Model.} The unified understanding and generation model \cite{xie2024muse} unifies token representations across different modalities, simultaneously modeling capabilities for understanding, generation, and editing. This enables the model to continually learn the deep connections between visual and textual information during training. In complex marketing scenarios, it achieves more intelligent, flexible, and interactive execution methods.


\section{Acknowledgements}
We sincerely appreciate Ya Li, Xibin Wu, Yixing Wu, Yingying Li, Yuzhong Wang, Liang Ma, Tuoyu Zhang, Shixiang Zhu for their valuable supports for our Aquarius project.

\bibliography{references}

\begin{thebibliography}{45}
\providecommand{\natexlab}[1]{#1}
\providecommand{\url}[1]{\texttt{#1}}
\expandafter\ifx\csname urlstyle\endcsname\relax
  \providecommand{\doi}[1]{doi: #1}\else
  \providecommand{\doi}{doi: \begingroup \urlstyle{rm}\Url}\fi

\bibitem[Bar-Tal et~al.(2023)Bar-Tal, Yariv, Lipman, and Dekel]{bartal2023multidiffusionfusingdiffusionpaths}
Omer Bar-Tal, Lior Yariv, Yaron Lipman, and Tali Dekel.
\newblock Multidiffusion: Fusing diffusion paths for controlled image generation, 2023.
\newblock URL \url{https://arxiv.org/abs/2302.08113}.

\bibitem[BlackForestLabs(2024)]{flux2024}
BlackForestLabs.
\newblock Flux.
\newblock \url{https://github.com/black-forest-labs/flux}, 2024.

\bibitem[Brooks et~al.(2024)Brooks, Peebles, Holmes, DePue, Guo, Jing, Schnurr, Taylor, Luhman, Luhman, Ng, Wang, and Ramesh]{videoworldsimulators2024}
Tim Brooks, Bill Peebles, Connor Holmes, Will DePue, Yufei Guo, Li~Jing, David Schnurr, Joe Taylor, Troy Luhman, Eric Luhman, Clarence Ng, Ricky Wang, and Aditya Ramesh.
\newblock Video generation models as world simulators.
\newblock 2024.
\newblock URL \url{https://openai.com/research/video-generation-models-as-world-simulators}.

\bibitem[Chen et~al.(2023)Chen, Yu, Ge, Yao, Xie, Wu, Wang, Kwok, Luo, Lu, and Li]{chen2023pixartalphafasttrainingdiffusion}
Junsong Chen, Jincheng Yu, Chongjian Ge, Lewei Yao, Enze Xie, Yue Wu, Zhongdao Wang, James Kwok, Ping Luo, Huchuan Lu, and Zhenguo Li.
\newblock Pixart-$\alpha$: Fast training of diffusion transformer for photorealistic text-to-image synthesis, 2023.
\newblock URL \url{https://arxiv.org/abs/2310.00426}.

\bibitem[Chen et~al.(2024)Chen, Shen, Ye, Cao, Tu, Bouganis, Zhao, and Chen]{chen2024deltadittrainingfreeaccelerationmethod}
Pengtao Chen, Mingzhu Shen, Peng Ye, Jianjian Cao, Chongjun Tu, Christos-Savvas Bouganis, Yiren Zhao, and Tao Chen.
\newblock $\delta$-dit: A training-free acceleration method tailored for diffusion transformers, 2024.
\newblock URL \url{https://arxiv.org/abs/2406.01125}.

\bibitem[Dai et~al.(2024)Dai, Deng, Zhao, Xu, Gao, Chen, Li, Zeng, Yu, Wu, et~al.]{dai2024deepseekmoe}
Damai Dai, Chengqi Deng, Chenggang Zhao, RX~Xu, Huazuo Gao, Deli Chen, Jiashi Li, Wangding Zeng, Xingkai Yu, Yu~Wu, et~al.
\newblock Deepseekmoe: Towards ultimate expert specialization in mixture-of-experts language models.
\newblock \emph{arXiv preprint arXiv:2401.06066}, 2024.

\bibitem[Dao et~al.()Dao, Fu, Ermon, Rudra, and R{\'e}]{dao2205fast}
T~Dao, DY~Fu, S~Ermon, A~Rudra, and C~Flashattention R{\'e}.
\newblock Fast and memory-efficient exact attention with io-awareness, 2022.
\newblock \emph{URL https://arxiv. org/abs/2205.14135}.

\bibitem[Esser et~al.(2024)Esser, Kulal, Blattmann, Entezari, Müller, Saini, Levi, Lorenz, Sauer, Boesel, Podell, Dockhorn, English, Lacey, Goodwin, Marek, and Rombach]{esser2024scalingrectifiedflowtransformers}
Patrick Esser, Sumith Kulal, Andreas Blattmann, Rahim Entezari, Jonas Müller, Harry Saini, Yam Levi, Dominik Lorenz, Axel Sauer, Frederic Boesel, Dustin Podell, Tim Dockhorn, Zion English, Kyle Lacey, Alex Goodwin, Yannik Marek, and Robin Rombach.
\newblock Scaling rectified flow transformers for high-resolution image synthesis, 2024.
\newblock URL \url{https://arxiv.org/abs/2403.03206}.

\bibitem[FFmpeg()]{Ffmpeg}
FFmpeg.
\newblock Ffmpeg.
\newblock URL \url{https://ffmpeg.org/}.

\bibitem[Goodfellow et~al.(2014)Goodfellow, Pouget-Abadie, Mirza, Xu, Warde-Farley, Ozair, Courville, and Bengio]{goodfellow2014generativeadversarialnetworks}
Ian~J. Goodfellow, Jean Pouget-Abadie, Mehdi Mirza, Bing Xu, David Warde-Farley, Sherjil Ozair, Aaron Courville, and Yoshua Bengio.
\newblock Generative adversarial networks, 2014.
\newblock URL \url{https://arxiv.org/abs/1406.2661}.

\bibitem[Grattafiori et~al.(2024)Grattafiori, Dubey, Jauhri, Pandey, Kadian, Al-Dahle, Letman, Mathur, Schelten, Vaughan, et~al.]{grattafiori2024llama}
Aaron Grattafiori, Abhimanyu Dubey, Abhinav Jauhri, Abhinav Pandey, Abhishek Kadian, Ahmad Al-Dahle, Aiesha Letman, Akhil Mathur, Alan Schelten, Alex Vaughan, et~al.
\newblock The llama 3 herd of models.
\newblock \emph{arXiv preprint arXiv:2407.21783}, 2024.

\bibitem[Ho and Salimans(2022)]{ho2022classifierfreediffusionguidance}
Jonathan Ho and Tim Salimans.
\newblock Classifier-free diffusion guidance, 2022.
\newblock URL \url{https://arxiv.org/abs/2207.12598}.

\bibitem[Ho et~al.(2020)Ho, Jain, and Abbeel]{ho2020denoisingdiffusionprobabilisticmodels}
Jonathan Ho, Ajay Jain, and Pieter Abbeel.
\newblock Denoising diffusion probabilistic models, 2020.
\newblock URL \url{https://arxiv.org/abs/2006.11239}.

\bibitem[Jacobs et~al.(1991)Jacobs, Jordan, Nowlan, and Hinton]{moe}
Robert~A. Jacobs, Michael~I. Jordan, Steven~J. Nowlan, and Geoffrey~E. Hinton.
\newblock Adaptive mixtures of local experts.
\newblock \emph{Neural Computation}, 3\penalty0 (1):\penalty0 79--87, 1991.
\newblock \doi{10.1162/neco.1991.3.1.79}.

\bibitem[Jacobs et~al.(2023)Jacobs, Tanaka, Zhang, Zhang, Song, Rajbhandari, and He]{jacobs2023deepspeed}
Sam~Ade Jacobs, Masahiro Tanaka, Chengming Zhang, Minjia Zhang, Shuaiwen~Leon Song, Samyam Rajbhandari, and Yuxiong He.
\newblock Deepspeed ulysses: System optimizations for enabling training of extreme long sequence transformer models.
\newblock \emph{arXiv preprint arXiv:2309.14509}, 2023.

\bibitem[Jiang et~al.(2024)Jiang, Lin, Zhong, Huang, Chen, Zhang, Peng, Li, Xie, Nong, et~al.]{jiang2024megascale}
Ziheng Jiang, Haibin Lin, Yinmin Zhong, Qi~Huang, Yangrui Chen, Zhi Zhang, Yanghua Peng, Xiang Li, Cong Xie, Shibiao Nong, et~al.
\newblock $\{$MegaScale$\}$: Scaling large language model training to more than 10,000 $\{$GPUs$\}$.
\newblock In \emph{21st USENIX Symposium on Networked Systems Design and Implementation (NSDI 24)}, pages 745--760, 2024.

\bibitem[Kong et~al.(2024)Kong, Tian, Zhang, Min, Dai, Zhou, Xiong, Li, Wu, Zhang, Wu, Lin, Yuan, Long, Wang, Wang, Li, Huang, Yang, Tan, Wang, Song, Bai, Wu, Xue, Wang, Wang, Liu, Li, Li, Wang, Yu, Deng, Li, Chen, Cui, Peng, Yu, He, Xu, Zhou, Xu, Tao, Lu, Liu, Zhou, Wang, Yang, Wang, Liu, Jiang, and Zhong]{kong2025hunyuanvideosystematicframeworklarge}
Weijie Kong, Qi~Tian, Zijian Zhang, Rox Min, Zuozhuo Dai, Jin Zhou, Jiangfeng Xiong, Xin Li, Bo~Wu, Jianwei Zhang, Kathrina Wu, Qin Lin, Junkun Yuan, Yanxin Long, Aladdin Wang, Andong Wang, Changlin Li, Duojun Huang, Fang Yang, Hao Tan, Hongmei Wang, Jacob Song, Jiawang Bai, Jianbing Wu, Jinbao Xue, Joey Wang, Kai Wang, Mengyang Liu, Pengyu Li, Shuai Li, Weiyan Wang, Wenqing Yu, Xinchi Deng, Yang Li, Yi~Chen, Yutao Cui, Yuanbo Peng, Zhentao Yu, Zhiyu He, Zhiyong Xu, Zixiang Zhou, Zunnan Xu, Yangyu Tao, Qinglin Lu, Songtao Liu, Dax Zhou, Hongfa Wang, Yong Yang, Di~Wang, Yuhong Liu, Jie Jiang, and Caesar Zhong.
\newblock Hunyuanvideo: A systematic framework for large video generative models, 2024.
\newblock URL \url{https://arxiv.org/abs/2412.03603}.

\bibitem[Korthikanti et~al.(2023)Korthikanti, Casper, Lym, McAfee, Andersch, Shoeybi, and Catanzaro]{korthikanti2023reducing}
Vijay~Anand Korthikanti, Jared Casper, Sangkug Lym, Lawrence McAfee, Michael Andersch, Mohammad Shoeybi, and Bryan Catanzaro.
\newblock Reducing activation recomputation in large transformer models.
\newblock \emph{Proceedings of Machine Learning and Systems}, 5:\penalty0 341--353, 2023.

\bibitem[Lin et~al.(2024)Lin, Ge, Cheng, Li, Zhu, Wang, He, Ye, Yuan, Chen, Jia, Zhang, Tang, Pang, She, Yan, Hu, Dong, Chen, Pan, Zhou, Dong, Tian, and Yuan]{lin2024opensoraplanopensourcelarge}
Bin Lin, Yunyang Ge, Xinhua Cheng, Zongjian Li, Bin Zhu, Shaodong Wang, Xianyi He, Yang Ye, Shenghai Yuan, Liuhan Chen, Tanghui Jia, Junwu Zhang, Zhenyu Tang, Yatian Pang, Bin She, Cen Yan, Zhiheng Hu, Xiaoyi Dong, Lin Chen, Zhang Pan, Xing Zhou, Shaoling Dong, Yonghong Tian, and Li~Yuan.
\newblock Open-sora plan: Open-source large video generation model, 2024.
\newblock URL \url{https://arxiv.org/abs/2412.00131}.

\bibitem[Lipman et~al.(2023)Lipman, Chen, Ben-Hamu, Nickel, and Le]{lipman2023flowmatchinggenerativemodeling}
Yaron Lipman, Ricky T.~Q. Chen, Heli Ben-Hamu, Maximilian Nickel, and Matt Le.
\newblock Flow matching for generative modeling, 2023.
\newblock URL \url{https://arxiv.org/abs/2210.02747}.

\bibitem[Luma(2024)]{Luma}
Luma.
\newblock Dream machine.
\newblock 2024.
\newblock URL \url{https://lumalabs.ai/dream-machine}.

\bibitem[Moritz et~al.(2018)Moritz, Nishihara, Wang, Tumanov, Liaw, Liang, Elibol, Yang, Paul, Jordan, and Stoica]{moritz2018raydistributedframeworkemerging}
Philipp Moritz, Robert Nishihara, Stephanie Wang, Alexey Tumanov, Richard Liaw, Eric Liang, Melih Elibol, Zongheng Yang, William Paul, Michael~I. Jordan, and Ion Stoica.
\newblock Ray: A distributed framework for emerging ai applications, 2018.
\newblock URL \url{https://arxiv.org/abs/1712.05889}.

\bibitem[Narayanan et~al.(2021)Narayanan, Shoeybi, Casper, LeGresley, Patwary, Korthikanti, Vainbrand, Kashinkunti, Bernauer, Catanzaro, et~al.]{narayanan2021efficient}
Deepak Narayanan, Mohammad Shoeybi, Jared Casper, Patrick LeGresley, Mostofa Patwary, Vijay Korthikanti, Dmitri Vainbrand, Prethvi Kashinkunti, Julie Bernauer, Bryan Catanzaro, et~al.
\newblock Efficient large-scale language model training on gpu clusters using megatron-lm.
\newblock In \emph{Proceedings of the international conference for high performance computing, networking, storage and analysis}, pages 1--15, 2021.

\bibitem[Peebles and Xie(2023)]{peebles2023scalablediffusionmodelstransformers}
William Peebles and Saining Xie.
\newblock Scalable diffusion models with transformers, 2023.
\newblock URL \url{https://arxiv.org/abs/2212.09748}.

\bibitem[Peng et~al.(2025)Peng, Zheng, Shen, Young, Guo, Wang, Xu, Liu, Jiang, Li, Wang, Ye, Ren, Ma, Liang, Lian, Wu, Zhong, Li, Gong, Lei, Cheng, Zhang, Li, Zhang, Hu, Huang, Wang, Zhao, Wang, Wei, and You]{peng2025opensora20trainingcommerciallevel}
Xiangyu Peng, Zangwei Zheng, Chenhui Shen, Tom Young, Xinying Guo, Binluo Wang, Hang Xu, Hongxin Liu, Mingyan Jiang, Wenjun Li, Yuhui Wang, Anbang Ye, Gang Ren, Qianran Ma, Wanying Liang, Xiang Lian, Xiwen Wu, Yuting Zhong, Zhuangyan Li, Chaoyu Gong, Guojun Lei, Leijun Cheng, Limin Zhang, Minghao Li, Ruijie Zhang, Silan Hu, Shijie Huang, Xiaokang Wang, Yuanheng Zhao, Yuqi Wang, Ziang Wei, and Yang You.
\newblock Open-sora 2.0: Training a commercial-level video generation model in 200k, 2025.
\newblock URL \url{https://arxiv.org/abs/2503.09642}.

\bibitem[Podell et~al.(2023)Podell, English, Lacey, Blattmann, Dockhorn, Müller, Penna, and Rombach]{podell2023sdxlimprovinglatentdiffusion}
Dustin Podell, Zion English, Kyle Lacey, Andreas Blattmann, Tim Dockhorn, Jonas Müller, Joe Penna, and Robin Rombach.
\newblock Sdxl: Improving latent diffusion models for high-resolution image synthesis, 2023.
\newblock URL \url{https://arxiv.org/abs/2307.01952}.

\bibitem[PySceneDetect.(2024)]{PySceneDetect}
PySceneDetect.
\newblock Pyscenedetect.
\newblock 2024.
\newblock URL \url{https://www.scenedetect.com.}

\bibitem[Radford et~al.(2021)Radford, Kim, Hallacy, Ramesh, Goh, Agarwal, Sastry, Askell, Mishkin, Clark, Krueger, and Sutskever]{radford2021learningtransferablevisualmodels}
Alec Radford, Jong~Wook Kim, Chris Hallacy, Aditya Ramesh, Gabriel Goh, Sandhini Agarwal, Girish Sastry, Amanda Askell, Pamela Mishkin, Jack Clark, Gretchen Krueger, and Ilya Sutskever.
\newblock Learning transferable visual models from natural language supervision, 2021.
\newblock URL \url{https://arxiv.org/abs/2103.00020}.

\bibitem[Rajbhandari et~al.(2020)Rajbhandari, Rasley, Ruwase, and He]{rajbhandari2020zero}
Samyam Rajbhandari, Jeff Rasley, Olatunji Ruwase, and Yuxiong He.
\newblock Zero: Memory optimizations toward training trillion parameter models.
\newblock In \emph{SC20: International Conference for High Performance Computing, Networking, Storage and Analysis}, pages 1--16. IEEE, 2020.

\bibitem[Ren et~al.(2024)Ren, Xia, Lu, Zhang, Wu, Xie, Wang, and Xiao]{ren2024hypersdtrajectorysegmentedconsistency}
Yuxi Ren, Xin Xia, Yanzuo Lu, Jiacheng Zhang, Jie Wu, Pan Xie, Xing Wang, and Xuefeng Xiao.
\newblock Hyper-sd: Trajectory segmented consistency model for efficient image synthesis, 2024.
\newblock URL \url{https://arxiv.org/abs/2404.13686}.

\bibitem[Rhu et~al.(2016)Rhu, Gimelshein, Clemons, Zulfiqar, and Keckler]{rhu2016vdnn}
Minsoo Rhu, Natalia Gimelshein, Jason Clemons, Arslan Zulfiqar, and Stephen~W Keckler.
\newblock vdnn: Virtualized deep neural networks for scalable, memory-efficient neural network design.
\newblock In \emph{2016 49th Annual IEEE/ACM International Symposium on Microarchitecture (MICRO)}, pages 1--13. IEEE, 2016.

\bibitem[Rombach et~al.(2022)Rombach, Blattmann, Lorenz, Esser, and Ommer]{rombach2022highresolutionimagesynthesislatent}
Robin Rombach, Andreas Blattmann, Dominik Lorenz, Patrick Esser, and Björn Ommer.
\newblock High-resolution image synthesis with latent diffusion models, 2022.
\newblock URL \url{https://arxiv.org/abs/2112.10752}.

\bibitem[Runway(2024)]{Runway}
Runway.
\newblock Runway-gen-3-alpha.
\newblock 2024.
\newblock URL \url{https://runwayml.com/research/introducing-gen-3-alpha}.

\bibitem[Shoeybi et~al.(2019)Shoeybi, Patwary, Puri, LeGresley, Casper, and Catanzaro]{shoeybi2019megatron}
Mohammad Shoeybi, Mostofa Patwary, Raul Puri, Patrick LeGresley, Jared Casper, and Bryan Catanzaro.
\newblock Megatron-lm: Training multi-billion parameter language models using model parallelism.
\newblock \emph{arXiv preprint arXiv:1909.08053}, 2019.

\bibitem[Sou{\v{c}}ek and Loko{\v{c}}(2020)]{soucek2020transnetv2}
Tom{\'a}{\v{s}} Sou{\v{c}}ek and Jakub Loko{\v{c}}.
\newblock Transnet v2: An effective deep network architecture for fast shot transition detection.
\newblock \emph{arXiv preprint arXiv:2008.04838}, 2020.

\bibitem[Su et~al.(2023)Su, Lu, Pan, Murtadha, Wen, and Liu]{su2023roformerenhancedtransformerrotary}
Jianlin Su, Yu~Lu, Shengfeng Pan, Ahmed Murtadha, Bo~Wen, and Yunfeng Liu.
\newblock Roformer: Enhanced transformer with rotary position embedding, 2023.
\newblock URL \url{https://arxiv.org/abs/2104.09864}.

\bibitem[WanTeam et~al.(2025)WanTeam, Wang, Ai, Wen, Mao, Xie, Chen, Yu, Zhao, Yang, Zeng, Wang, Zhang, Zhou, Wang, Chen, Zhu, Zhao, Yan, Huang, Feng, Zhang, Li, Wu, Chu, Feng, Zhang, Sun, Fang, Wang, Gui, Weng, Shen, Lin, Wang, Wang, Zhou, Wang, Shen, Yu, Shi, Huang, Xu, Kou, Lv, Li, Liu, Wang, Zhang, Huang, Li, Wu, Liu, Pan, Zheng, Hong, Shi, Feng, Jiang, Han, Wu, and Liu]{wan2025wanopenadvancedlargescale}
WanTeam, Ang Wang, Baole Ai, Bin Wen, Chaojie Mao, Chen-Wei Xie, Di~Chen, Feiwu Yu, Haiming Zhao, Jianxiao Yang, Jianyuan Zeng, Jiayu Wang, Jingfeng Zhang, Jingren Zhou, Jinkai Wang, Jixuan Chen, Kai Zhu, Kang Zhao, Keyu Yan, Lianghua Huang, Mengyang Feng, Ningyi Zhang, Pandeng Li, Pingyu Wu, Ruihang Chu, Ruili Feng, Shiwei Zhang, Siyang Sun, Tao Fang, Tianxing Wang, Tianyi Gui, Tingyu Weng, Tong Shen, Wei Lin, Wei Wang, Wei Wang, Wenmeng Zhou, Wente Wang, Wenting Shen, Wenyuan Yu, Xianzhong Shi, Xiaoming Huang, Xin Xu, Yan Kou, Yangyu Lv, Yifei Li, Yijing Liu, Yiming Wang, Yingya Zhang, Yitong Huang, Yong Li, You Wu, Yu~Liu, Yulin Pan, Yun Zheng, Yuntao Hong, Yupeng Shi, Yutong Feng, Zeyinzi Jiang, Zhen Han, Zhi-Fan Wu, and Ziyu Liu.
\newblock Wan: Open and advanced large-scale video generative models, 2025.
\newblock URL \url{https://arxiv.org/abs/2503.20314}.

\bibitem[Xie et~al.(2024)Xie, Du, Song, and Liu]{xie2024muse}
Rongchang Xie, Chen Du, Ping Song, and Chang Liu.
\newblock Muse-vl: Modeling unified vlm through semantic discrete encoding.
\newblock \emph{arXiv preprint arXiv:2411.17762}, 2024.

\bibitem[Xu et~al.(2024)Xu, Zou, Huang, Chen, Liu, Cheng, Shi, and Huang]{xu2024easyanimatehighperformancelongvideo}
Jiaqi Xu, Xinyi Zou, Kunzhe Huang, Yunkuo Chen, Bo~Liu, MengLi Cheng, Xing Shi, and Jun Huang.
\newblock Easyanimate: A high-performance long video generation method based on transformer architecture, 2024.
\newblock URL \url{https://arxiv.org/abs/2405.18991}.

\bibitem[Xue et~al.(2021)Xue, Constant, Roberts, Kale, Al-Rfou, Siddhant, Barua, and Raffel]{xue2021mt5massivelymultilingualpretrained}
Linting Xue, Noah Constant, Adam Roberts, Mihir Kale, Rami Al-Rfou, Aditya Siddhant, Aditya Barua, and Colin Raffel.
\newblock mt5: A massively multilingual pre-trained text-to-text transformer, 2021.
\newblock URL \url{https://arxiv.org/abs/2010.11934}.

\bibitem[Yang et~al.(2024)Yang, Teng, Zheng, Ding, Huang, Xu, Yang, Hong, Zhang, Feng, Yin, Zhang, Wang, Cheng, Xu, Gu, Dong, and Tang]{yang2025cogvideoxtexttovideodiffusionmodels}
Zhuoyi Yang, Jiayan Teng, Wendi Zheng, Ming Ding, Shiyu Huang, Jiazheng Xu, Yuanming Yang, Wenyi Hong, Xiaohan Zhang, Guanyu Feng, Da~Yin, Yuxuan Zhang, Weihan Wang, Yean Cheng, Bin Xu, Xiaotao Gu, Yuxiao Dong, and Jie Tang.
\newblock Cogvideox: Text-to-video diffusion models with an expert transformer, 2024.
\newblock URL \url{https://arxiv.org/abs/2408.06072}.

\bibitem[Yu et~al.(2024)Yu, Lezama, Gundavarapu, Versari, Sohn, Minnen, Cheng, Birodkar, Gupta, Gu, Hauptmann, Gong, Yang, Essa, Ross, and Jiang]{yu2024languagemodelbeatsdiffusion}
Lijun Yu, José Lezama, Nitesh~B. Gundavarapu, Luca Versari, Kihyuk Sohn, David Minnen, Yong Cheng, Vighnesh Birodkar, Agrim Gupta, Xiuye Gu, Alexander~G. Hauptmann, Boqing Gong, Ming-Hsuan Yang, Irfan Essa, David~A. Ross, and Lu~Jiang.
\newblock Language model beats diffusion -- tokenizer is key to visual generation, 2024.
\newblock URL \url{https://arxiv.org/abs/2310.05737}.

\bibitem[Zhao et~al.(2025)Zhao, Jin, Wang, and You]{zhao2025realtimevideogenerationpyramid}
Xuanlei Zhao, Xiaolong Jin, Kai Wang, and Yang You.
\newblock Real-time video generation with pyramid attention broadcast, 2025.
\newblock URL \url{https://arxiv.org/abs/2408.12588}.

\bibitem[Zhou et~al.(2023)Zhou, Li, Chan, and Loy]{zhou2023propainter}
Shangchen Zhou, Chongyi Li, Kelvin~C.K Chan, and Chen~Change Loy.
\newblock {ProPainter}: Improving propagation and transformer for video inpainting.
\newblock In \emph{Proceedings of IEEE International Conference on Computer Vision (ICCV)}, 2023.

\bibitem[Zhu et~al.(2024)Zhu, Feng, Lu, Fang, and Yang]{zhu2024zerof}
Jian Zhu, Peicheng Feng, Jiawei Lu, Bowei Fang, and Hesong Yang.
\newblock Zerof-offload: forward-gradient scheme for efficient full parameter fine-tuning of billion-scale language models.
\newblock \emph{Machine Learning: Science and Technology}, 5\penalty0 (4):\penalty0 045054, 2024.

\end{thebibliography}

\end{document}